\documentclass{article}
\usepackage[preprint]{neurips_2024}
\usepackage{natbib}
\setcitestyle{numbers,square}

\usepackage{amsmath,amsfonts,bm}

\def\eqref#1{equation~\ref{#1}}

\def\1{\bm{1}}

\DeclareMathAlphabet{\mathsfit}{\encodingdefault}{\sfdefault}{m}{sl}
\SetMathAlphabet{\mathsfit}{bold}{\encodingdefault}{\sfdefault}{bx}{n}

\usepackage{amsthm}
\usepackage{hyperref}
\usepackage{url}
\usepackage{graphicx} 
\usepackage{bbm}
\newtheorem{theorem}{Theorem}[section]
\newtheorem{corollary}{Corollary}[theorem]

\newtheorem{assumption}[theorem]{Assumption}
\newtheorem{proposition}[theorem]{Proposition}
\usepackage[ruled,vlined, linesnumbered, boxed,norelsize]{algorithm2e}
\usepackage{tikz}
\usepackage{wrapfig}
\usepackage{bbm}
\usepackage{subcaption}
\usepackage{colortbl}
\usepackage{booktabs}
\usepackage{enumitem}
\usepackage{graphicx}
\usepackage{wrapfig}
\usepackage{subcaption}
\usepackage[ruled,vlined, linesnumbered, boxed,norelsize]{algorithm2e}
\usepackage{amsmath}
\usepackage{lipsum}
\usepackage{mdframed}
\usepackage{hyperref}
\usepackage{booktabs}

\newmdenv[
  linecolor=gray,
  outerlinewidth=1pt,
  roundcorner=5pt,
  innertopmargin=0.5\baselineskip,
  innerbottommargin=0.5\baselineskip,
  innerrightmargin=10pt,
  innerleftmargin=10pt,
  backgroundcolor=gray!10,
  outerlinecolor=gray,
  frametitlebackgroundcolor=gray!20,
  frametitlerule=true,
  frametitlerulecolor=gray,
  skipabove=\baselineskip,
  skipbelow=\baselineskip
]{promptBox}

\def\ty{\tilde{y}}

\def\na{\text{naive}}

\title{Direct Alignment of Language Models \\
via Quality-Aware Self-Refinement}

\author{
Runsheng Yu$^1$\thanks{Work done during the internship at Tencent LightSpeed Studios.} ~~~Yong Wang$^2$~~~Xiaoqi Jiao$^2$~~~Youzhi Zhang$^3$~~~James T. Kwok$^1$ \\
\small $^1$Hong Kong University of Science and Technology \\
\small $^2$LightSpeed Studios, Tencent \\
\small $^3$Centre for Artificial Intelligence and Robotics, Hong Kong Institute of Science \& Innovation, CAS  \\
\small \texttt{\{runshengyu,seaywang\}@gmail.com,xiaoqijiao@tencent.com}\\
\small \texttt{youzhi.zhang@cair-cas.org.hk,jamesk@cse.ust.hk}
}

\begin{document}
\maketitle
\begin{abstract}
Reinforcement Learning from Human Feedback (RLHF) has been commonly used to align the behaviors of Large Language Models (LLMs) with human preferences. Recently, a popular alternative is Direct Policy Optimization (DPO), which replaces an LLM-based reward model with the policy itself, thus obviating the need for extra memory and training time to learn the reward model. However, DPO does not consider the relative qualities of the positive and negative responses, and can lead to sub-optimal training outcomes. To alleviate this problem, we investigate the use of intrinsic knowledge within the on-the-fly fine-tuning LLM to obtain relative qualities and help to refine the loss function. Specifically, we leverage the knowledge of the LLM to design a refinement function to estimate the quality of both the positive and negative responses. We show that the constructed refinement function can help self-refine the loss function under mild assumptions. The refinement function is integrated into DPO and its variant Identity Policy Optimization (IPO). Experiments across various evaluators indicate that they can improve the performance of the fine-tuned models over DPO and IPO.
\end{abstract}

\section{Introduction}
Large Language Models (LLMs) have demonstrated significant capabilities across various natural language processing tasks \citep{radford2019language,zhang2022opt, touvron2023llama, achiam2023gpt}. Ensuring that these LLMs produce the desired responses and behaviors that are aligned with human preferences is crucial for safe and controllable AI systems \citep{ouyang2022training}. To achieve this, a popular method is Reinforcement Learning from Human Feedback (RLHF) \citep{christiano2017deep, achiam2023gpt, bai2022training}, which first trains a reward model using human-labeled response pairs, and then uses this to adjust the policy parameters of the LLM \citep{bai2022training, ouyang2022training}. However, the reward model is often constructed by another LLM, which requires further training and storage \citep{amini2024direct}.

To reduce the storage and training time of the reward model, a variety of methods have been proposed recently \citep{dong2023raft, yuan2023rrhf, amini2024direct}. In particular, a prominent solution is the Direct Policy Optimization (DPO) \citep{amini2024direct}, which replaces the reward model with the policy itself, thus obviating the need for an explicit reward model. Recently, numerous variants of DPO have also been developed \citep{wang2023beyond, azar2024general, amini2024direct, ethayarajh2024kto, song2024icdpo}.

The objective of DPO is to consistently increase the likelihood of human-preferred responses while reducing the likelihood of the undesired ones. However, this strategy does not consider the relative qualities of the positive and negative responses, and can lead to suboptimal training outcomes, particularly when the preferred responses are not substantially superior, or when the undesired responses are not adequately inferior \citep{amini2024direct, tunstall2023zephyr, cui2023ultrafeedback}. 

To alleviate this issue, Amini et al. \citep{amini2024direct} and Zhou et al. \citep{zhou2023beyond} propose the use of a score function to self-refine the objective. However, this approach requires the availability of an ideal reward or score function, which may not be always feasible. Similarly, Cui et al. \citep{cui2023ultrafeedback} and Tunstall et al. \cite{tunstall2023zephyr} employ GPT-4 \citep{achiam2023gpt} to select high-quality response pairs by scoring them. This method requires a strong LLM to effectively filter the dataset, which again may not always be practical. These considerations raise the question: \emph{Can we achieve this by using the inherent knowledge within the policy itself?}

Recently, self-alignment has attracted increasing attention due to its ability to leverage the inherent knowledge of LLMs to enhance alignment capabilities, obviating the necessity for additional human-annotated data \citep{munos2023nash, alami2024investigating, lee2024aligning, yuan2024self}. Inspired by this, we propose to utilize on-the-fly fine-tuning of an LLM's knowledge to help evaluate the quality of positive and negative responses. The underlying premise is that even relatively weak LLMs possess some ability to assess the quality of responses \citep{ji2023ai, ji2024aligner}. Consequently, our objective is to exploit this capability to more effectively evaluate response quality, thereby enhancing the efficiency and accuracy of the fine-tuned model.

In this paper, we investigate the use of intrinsic knowledge within the LLM to self-refine the loss function. In summary, the contributions of this work are as follows:
\begin{itemize}[leftmargin=*]
\item We leverage the knowledge of the LLM to design a refinement function, which estimates the quality of positive and negative responses.
\item We demonstrate that the constructed refinement function can help self-refine the loss function under mild assumptions.
\item By utilizing the refinement function, we propose two novel approaches based on DPO and its variant Identity Policy Optimization (IPO)
\citep{azar2024general}.
\end{itemize}
Experimental results across various evaluators indicate that the proposed self-refined methods improve the performance of the fine-tuned models compared to their counterparts.

\section{Preliminaries}
\subsection{Classical RLHF with Bradley-Terry Reward Model}
Given a pre-trained large language model (LLM) $\pi_{\text{ref}}$ as initialization, Reinforcement Learning from Human Feedback (RLHF) \citep{christiano2017deep, achiam2023gpt, bai2022training} aims to learn an LLM $\pi$ that aligns with human values and preferences. Specifically, let $x$ be the query, $y$ be the output of $\pi$, and $r$ be a reward function that evaluates the performance of $y$ given $x$, RLHF tries to maximize $r$ while ensuring that the trained LLM $\pi$ does not deviate significantly from the pre-trained model $\pi_{\text{ref}}$. This can be formulated as the following optimization problem \citep{ouyang2022training, nika2024reward, achiam2023gpt}:
\begin{equation}\label{eq:rl_obj}
\max_{\pi} \mathbb{E}_{x \sim \mathcal{D}, y \sim \pi(\cdot | x)}r(y | x) - \beta \mathbb{D}_{\mathrm{KL}}[\pi(y | x) \| \pi_{\text{ref}}(y | x)],
\end{equation}
where $\mathbb{D}_{\mathrm{KL}}[\pi(y | x) \| \pi_{\text{ref}}(y | x)]$ is the Kullback-Leibler divergence between $\pi$ and $\pi_{\text{ref}}$, and $\beta$ is a constant.

Since $r$ is unknown, the user needs to provide a set of preferences $D \equiv \{(x_i, y^+_i, y^-_i)\}^N_{i=1}$, where $y^+_i$ (resp. $y^-_i$) is the positive (resp. negative) response for query $x_i$. A suitable model, typically another LLM (parameterized by $\omega$) \citep{lambert2022illustrating,achiam2023gpt}, then learns the reward function by maximizing the probability $p(y_i^+ \succ y_i^-|x)$ that $y_i^+$ is preferred over $y_i^-$ (denoted $y_i^+ \succ y_i^-$) \citep{ouyang2022training,amini2024direct,azar2024general}. Typically, this probability is defined by the Bradley-Terry preference model \citep{bradley1952rank, christiano2017deep} as:
\begin{equation} \label{eq:bt}
p(y_i^+ \succ y_i^- | x_i) \equiv \sigma(r(y_i^+ | x_i) - r(y_i^- | x_i)),
\end{equation}
where $\sigma(\cdot)$ is the logistic function. The optimal $r$ can then be obtained by maximizing the log likelihood 
\begin{equation}\label{eq:train_reward}
\max_r \mathbb{E}_{(x, y^+, y^-) \sim \mathcal{D}} \log 
[\sigma(r(y_i^+ | x_i) - r(y_i^- | x_i))].
\end{equation}
With the obtained $r$, we can then find $\pi$ by optimizing (\ref{eq:rl_obj}).

\subsection{Direct Preference Optimization (DPO)}
In RLHF, the reward model is represented by a LLM \citep{ouyang2022training}. This can be time- and memory-expensive. It is observed that the optimal $\pi$ in (\ref{eq:rl_obj}) indeed has the closed form \citep{rafailov2023direct}:
$\pi(y |x) \propto \pi_{\text{ref}}(y | x) \exp\left(\frac{r(y | x)}{\beta}\right),$ and so
\begin{equation} \label{eq:dpo}
r(y | x) = \beta \log \frac{\pi(y | x)}{\pi_{\text{ref}}(y | x)}+c(x),
\end{equation}
where $c(x) \equiv \sum_{y \in \mathcal{Y}} \pi_{\text{ref}}(y \mid x) \exp \left(\frac{1}{\beta} r(y|x)\right)$, and $\mathcal{Y}$ is the set of responses. By plugging this into (\ref{eq:train_reward}), the optimal policy can be found directly by maximizing:
\begin{equation}
\label{eq:obj_dro}
\max\nolimits_{\pi} \mathbb{E}_{(x, y^+, y^-) \sim \mathcal{D}}\log\sigma\left(\beta \log \frac{\pi(y^+ | x)}{\pi_{\text{ref}}(y^+ | x)} - \beta \log \frac{\pi(y^- | x)}{\pi_{\text{ref}}(y^- | x)} \right). 
\end{equation}

\subsection{Identity Policy Optimisation (IPO)}
DPO aims to maximize (\ref{eq:rl_obj}). However, our goal is to maximize the preference rather than the reward \citep{wirth2017survey}. Consequently, a better option is to maximize the preference probabilities $p(y \succ y' | x)$. To this end, IPO \citep{azar2024general} optimizes the following objective:
\begin{equation}\label{eq:ipo_obj}
\max_{\pi} 
\mathbb{E}_{x \sim \mathcal{D}, y^+ \sim \pi(\cdot | x), y^- \sim \pi_{\text{ref}}(\cdot
| x)} p(y^+ \succ y^- | x) -\beta
\mathbb{D}_{\mathrm{KL}}[\pi(y^+ | x) \| \pi_{\text{ref}}(y^- | x)].
\end{equation}

Similar to DPO, following \citep{azar2024general},  the optimal $\pi$ also has a closed-form solution:
\begin{equation}\label{eq:closed_form_ipo}
\pi(y^+ | x) \propto \pi_{\text{ref}}(y^- | x) \exp\left(\mathbb{E}_{y^+ \sim \pi_{\text{ref}}(\cdot | x)}p(y^+ \succ y^- | x) /\beta \right).
\end{equation}

The loss for IPO is then defined as:
\begin{equation} \label{eq:ipo-loss}
\mathbb{E}_{(x, y^+, y^-) \sim \mathcal{D}}\left[ \left(\log  \frac{\pi(y^+ | x)}{\pi_{\text{ref}}(y^+ | x)} - \log \frac{\pi(y^- | x)}{\pi_{\text{ref}}(y^- | x)} - \frac{1}{2\beta}\right)^2\right].
\end{equation}

\subsection{Self-Alignment} 
The proposed method is also related to self-alignment~\citep{lee2024aligning,yuan2024self,madaan2024self}, which uses the LLM's own knowledge to improve the alignment. For example, self-judgement \citep{lee2024aligning,yuan2024self,sun2024principle,madaan2024self,zhao2023survey} uses the LLM as a judge to evaluate the generated answer. Self-improving \citep{munos2023nash,alami2024investigating,cheng2024self,chen2024self,calandriello2024human} improves LLM by playing game with another LLM opponent. Our work uses the self-alignment ability of LLM to construct a refinement function to adjust the quality of responses.

\section{Proposed Method}
\subsection{Limitation of Maximizing Bradley-Terry Preference} \label{sec:limitation}
Let $r^*$ and $r$ be the true and learned reward function respectively. For any tuple $(x, y^+, y^-) \in D$, the true reward of the positive response $y^+$ should exceed that of the negative response $y^-$, i.e., $y^+ \succ y^- \Leftrightarrow r^*(y^+|x) > r^*(y^-|x)$. Consider two tuples $(x_i, y_i^+, y_i^-)$ and $(x_j, y_j^+, y_j^-)$. When $r^*(y_i^+|x_i) - r^*(y_i^-|x_i)>r^*(y_j^+|x_j) - r^*(y_j^-|x_j)$, we prefer $p(y_i^+ \succ y_i^- | x_i) > p(y_j^+ \succ y_j^- | x_j)$, and so $\sigma(r(y_i^+|x_i) - r(y_i^-|x_i))> \sigma(r(y_j^+|x_j) - r(y_j^-|x_j))$ from (\ref{eq:bt}). In other words, the more informative tuple $(x_i, y_i^+, y_i^-)$ should be more important. However, RLHF simply maximizes (\ref{eq:train_reward}), which considers all tuples in the dataset $D$ equally.

\subsection{Refining the Reward Difference between Positive and Negative Responses}
\subsubsection{Intuition}
\label{sec:intuition}
To alleviate this problem, we propose adding a {\em refinement} function $\Delta( y^-, y^+;x):\mathcal{Y} \times \mathcal{Y} \times \mathcal{X} \rightarrow \mathbb{R}$ (where $\mathcal{X}$ and $\mathcal{Y}$ are the sets of queries and responses, respectively) to adjust $r(y^+|x) - r(y^-|x)$, so that problem (\ref{eq:train_reward}) is modified to:
\begin{equation}\label{eq:bt_obj2}
\max_\pi  \mathbb{E}_{(x, y^+, y^-) \sim \mathcal{D}} \log[\sigma(r(y^+|x) - r(y^-|x)-\lambda \Delta( y^-, y^+;x))],
\end{equation}
where $\lambda$ is a positive constant. Intuitively, when $\Delta( y^-, y^+;x)$ is large, $\sigma(r(y^+|x) - r(y^-|x)-\lambda \Delta( y^-, y^+;x))$ becomes small, and the optimization in (\ref{eq:bt_obj2}) will tend to enlarge the “distance” $r(y^+|x) - r(y^-|x)$ between the positive and negative responses. Thus, for two tuples $(x, y^+, y^-)$ and $(x, \ty^{+}, \ty^{-})$ corresponding to the same query $x$, we want to design a $\Delta$ such that when $\Delta( y^-, y^+;x) > \Delta( \ty^-, \ty^+;x)$, their true reward values satisfy: \[ r^*(y^+|x) - r^*(y^-|x) > r^*(\ty^{+}|x) - r^*(\ty^{-}|x). \] In other words, a larger difference \( r^*(y^+|x) - r^*(y^-|x)  \) in the true reward values between the positive and negative responses corresponds to a larger $\Delta(y^-, y^+;x)$, and vice versa. However, obviously the difficulty is that we do not have access to $r^*$. 

\subsubsection{Implementing $\Delta$ via Prompting}
First, we assume that the LLM is capable of learning a reward function that aligns with the true reward function. This premise is formalized as follows:
\begin{assumption} 
\label{assu:improve}
LLM $\pi$ can construct a reward model $r$ such that for any $(x,y^+,y^-)$ with $y^+\succ y^-$, the corresponding reward values satisfy $r(y^+|x)>r(y^-|x)$.
\end{assumption}

As $r^*$ is unknown, with a capable LLM, a natural idea is to use $r$ as a proxy of $r^*$. Recall from Section~\ref{sec:intuition} that a large $r^*(y^+|x) - r^*(y^-|x)$ should correspond to a large $\Delta( y^-, y^+;x)$. Using (\ref{eq:dpo}), one can define $\Delta$ as 
\begin{equation} \label{eq:naive}
\Delta_{\na} = r(y^+|x) - r(y^-|x) =\beta\left(\log\frac{\pi (y^{+} |x)}{\pi_{\text{ref}} (y^{+} |x)} -\log \frac{\pi (y^{-} |x)}{\pi_{\text{ref}} (y^{-} |x)}\right). 
\end{equation}
Substituting this into the DPO objective (\ref{eq:obj_dro}), we have
\begin{eqnarray*}
\lefteqn{\mathbb{E}_{(x,y^{+} ,y^{-})\sim \mathcal{D}}
\log \sigma\left(\beta \log \frac{\pi(y^+ | x)}{\pi_{\text{ref}}(y^+ | x)} - \beta \log \frac{\pi(y^- | x)}{\pi_{\text{ref}}(y^- | x)} - \lambda 
\Delta_{\na}( y^-,y^+;x)
\right)}\\
& = & \mathbb{E}_{(x,y^{+} ,y^{-}) \sim \mathcal{D}}
\log 
\sigma\left((\beta-\lambda\beta) \log \frac{\pi(y^+ | x)}{\pi_{\text{ref}}(y^+ | x)} - (\beta-\lambda\beta) \log \frac{\pi(y^- | x)}{\pi_{\text{ref}}(y^- | x)}
\right). 
\end{eqnarray*}
However, this is the same as the original DPO objective except for a scaling of the regularization parameter $\beta$, and thus is not useful.

To alleviate this problem, our idea is to first improve the LLM performance by concatenating the query $x$ with a prompt $p$, as $p\oplus x$,
where $\oplus$ denotes concatenation. With (\ref{eq:dpo}) the reward with this prompt-augmented query is:
\begin{equation}\label{eq:constructed_r}
r( y|p\oplus x) =\beta \log \frac{\pi (y|p\oplus x)}{\pi_{\text{ref}} (y|p\oplus x)} +c(x).
\end{equation}

With a good prompt, this $r$ is expected to be a good proxy of $r^*$. Specifically, we use the following prompt $p$.
\begin{promptBox}
Please generate a response with a usefulness rating of 100 out of 100 for the following query. Note that the response should be harmless. The term 100 indicates the level of usefulness, where 100 is the maximum and 1 is the minimum. Query:
\end{promptBox}

We assume that adding this prompt does not change the preference between the positive and negative responses ($y^{+}$ and $y^{-}$). 
\begin{assumption} 
\label{assu:consistency} 
For a given query $x$, if $y^{+} \succ y^{-}$, we still have $y^{+}\succ y^{-}$ with the prompt-augmented query $p\oplus x$.
\end{assumption}

Analogous to (\ref{eq:naive}), we consider the following refinement function:
\begin{equation} \label{eq:delta}
 \Delta_\pi\left( y^-,y^+;x\right) \equiv \beta\log  \frac{\pi (y^+|p\oplus x)\pi_{\text{ref}} (y^- |p\oplus x)}{\pi_{\text{ref}} (y^+ |p\oplus x)\pi (y^- |p\oplus x)}.
\end{equation}

Note that we have added subscript $\pi$ to explicitly indicate the dependence of $\Delta$ on $\pi$. Obviously, when putting this $\Delta_\pi$ into the DPO objective (\ref{eq:obj_dro}), it does not suffer from the same problem as the $\Delta_{\na}$ discussed earlier.

The following Proposition shows that $\Delta_\pi$ in (\ref{eq:delta}) satisfies two important properties. First, $\Delta_\pi(y^-, y^+;x)$ can be measured relative to the optimal response $y^*$, which allows to represent $\Delta_\pi(y^-, y^+;x)$ in terms of $\Delta_\pi(y^-, y^*;x)$ and $\Delta_\pi(y^+, y^*;x)$. Second, the positive response (which has a higher true reward value) is ``closer" to $y^*$ than the negative response, and vice versa. All the proofs are in Appendix~\ref{sec:proofs}.

\begin{proposition}
\label{prop:displacement}
With Assumptions \ref{assu:improve} and \ref{assu:consistency}, we have
     (i) 
     $\Delta_\pi(y^-, y^+;x)= \Delta_\pi(y^-, y^*;x) - \Delta_\pi(y^+, y^*;x)$, where $y^*$ is the optimal $y$ for the given $x$;
     (ii)
    For any tuple $(x,y^+,y^-)$, $r^*(y^+|x) >r^*(y^-|x) \Leftrightarrow \Delta_\pi(y^+, y^*;x) < \Delta_\pi(y^-, y^*;x)$.
\end{proposition} 

The following Corollary shows that this $\Delta_\pi(y^-, y^+;x)$ satisfies the desired property in Section~\ref{sec:intuition}, namely that a larger difference \( r^*(y^+|x) - r^*(y^-|x)  \) in the true reward values between the positive and negative responses corresponds to a larger $\Delta_\pi(y^-, y^+;x)$.

\begin{corollary} 
\label{cor:relative_displacement}
For any $(x,y_i^+,y_i^-)$ and $(x,y_j^+,y_j^-)$, if $r^*(y_i^+|x)>r^*(y_j^+|x)$, and $r^*(y_i^-|x)<r^*(y_j^-|x)$ we have
$r^*(y_i^+|x) - r^*(y_i^-|x) > r^*(y_j^{+}|x) - r^*(y_j^{-}|x)
\Leftrightarrow 
\Delta_\pi(y_i^-, y_i^+;x)> \Delta_\pi(y_j^{-},y_j^{+};x)$.
\end{corollary} 

\subsection{Integration with DPO and IPO}
\subsubsection{Integration with DPO}
First, we show how to integrate the proposed refinement into DPO. Substituting (\ref{eq:dpo}) and (\ref{eq:delta}) into (\ref{eq:bt_obj2}), we obtain:
\begin{eqnarray}
\lefteqn{
\max_\pi \mathbb{E}_{(x,y^{+} ,y^{-}) \sim \mathcal{D}}\log
\sigma(r(y^+|x) - r(y^-|x) - \lambda 
\Delta_\pi( y^-,y^+;x)
)} \nonumber\\
&=& \!\!\!\!\!\!
\max_\pi \mathbb{E}_{(x,y^{+} ,y^{-})\sim \mathcal{D}}
\log \sigma\left(\beta \log \frac{\pi(y^+ | x)}{\pi_{\text{ref}}(y^+ | x)} \! - \! \beta \log \frac{\pi(y^- | x)}{\pi_{\text{ref}}(y^- | x)} - \lambda 
\Delta_\pi( y^-,y^+;x)
\right). \label{eq:Sr-DPO_obj}
\end{eqnarray}

Recall that $\Delta_\pi$ above depends on $\pi$. During learning, we use the stop-gradient operator\footnote{In other words, $\nabla_x\perp\!\![g(x)] \equiv 0$ and $\perp\!\![g(x)] \equiv g(x)$ for any differentiable $g$.} $\perp\!\![\cdot]$ on $\Delta_\pi$to prevent it from being changed. The whole procedure, which will be called Self-refined DPO (Sr-DPO), is shown in Algorithm \ref{alg:Sr-DPO}.
\SetKwComment{Comment}{/* }{ */}
\SetKwInput{KwInput}{Input}
\SetKwInput{KwOutput}{Output}

\begin{algorithm}[H]
\caption{Self-refined Direct Policy Optimization (Sr-DPO).}
\label{alg:Sr-DPO}
\KwInput{Dataset $D =\{(x,y^+,y^-)\}$, a pre-trained LLM $\pi_{\text{ref}}$ with parameter $\boldsymbol{\theta}_0$, learning rate $\alpha$.}
{
\For{$t = 1,2,\dots, T$}{
sample a minibatch $B$ of size $N$ from $D$;\\
$\boldsymbol{\theta}_t\!\! = \!\! \boldsymbol{\theta}_{t-1}+ \alpha\nabla_{\boldsymbol{\theta}} \frac{1}{N}\sum^N_{i=1}\!\log\sigma\left(\beta\left(\log\frac{\pi (y_i^{+} |x_i)}{\pi_{\text{ref}} (y_i^{+} |x_i)} \!- \! \log \frac{\pi (y_i^{-} |x_i)}{\pi_{\text{ref}} (y_i^{-} |x_i)}\right) \! - \! \lambda \!\perp\![\Delta_\pi( y_i^-,y_i^+;x_i)] \right)$;
}
}
\Return $\boldsymbol{\theta}_{T}$.
\end{algorithm}

\subsubsection{Integration with IPO }
For IPO, we first construct a variant of the IPO objective in (\ref{eq:ipo_obj}):
\begin{eqnarray}
\!\!\!\!\!\!\!\!\!\!\!\!\!\! & \max_\pi & \!\!\!\!\!\!\! \mathbb{E}_{x \sim \mathcal{D},  y^{+} \sim \pi(\cdot | x), y^{-} \sim \pi_{\text{ref}}(\cdot | x)} p(y^{+} \succ y^{-} | x) 
\!\! + \!\! \lambda 
r (y^{+} | p\oplus x) \!\! - \!\! \beta \mathbb{D}_{\mathrm{KL}}[\pi(y^{+} | x) \| \pi_{\text{ref}}(y^{-} | x)],
\label{eq:ipo_Cali}
\end{eqnarray}

which adds an extra $\lambda \mathbb{E}_{x\sim \mathcal{D} ,y^{+} \sim \pi (\cdot | x)} r (y^{+} | p\oplus x)$ term to maximize the expectation of reward $r (y^{+} | p\oplus x)$.

\begin{proposition} \label{prop:ipo_key}
The optimal policy $\pi$ in (\ref{eq:ipo_Cali}) satisfies
\begin{equation}\label{eq:ipo_eqv}
\log\frac{\pi (y^{+} |x)}{\pi_{\text{ref}} (y^{+} |x)} -\log \frac{\pi (y^{-} |x)}{\pi_{\text{ref}} (y^{-} |x)} =\frac{1}{2\beta } +\lambda \Delta_{\pi}( y^-,y^+;x).
\end{equation}
\end{proposition}

Note that when $\beta = 0$, $\log\frac{\pi (y^{+} |x)}{\pi_{\text{ref}} (y^{+} |x)} -\log \frac{\pi (y^{-} |x)}{\pi_{\text{ref}} (y^{-} |x)}-\lambda\Delta_\pi( y^-,y^+;x)$ should be infinite. This implies maximizing $\log\frac{\pi (y^{+} |x)}{\pi_{\text{ref}} (y^{+} |x)} -\log \frac{\pi (y^{-} |x)}{\pi_{\text{ref}} (y^{-} |x)}-\lambda\Delta_\pi( y^-,y^+;x)$ w.r.t, $\pi$, which has a similar form to DPO. The procedure, which will be called Self-Refined IPO (Sr-IPO), is shown in Algorithm \ref{alg:Sr-IPO}. It uses stochastic gradient descent to minimize the difference between the LHS and RHS in (\ref{eq:ipo_eqv}).

\begin{algorithm}[H]
\caption{Self-refined Identity Policy Optimization (Sr-IPO).}
\label{alg:Sr-IPO}
\KwInput{Dataset $D =\{(x,y^+,y^-)\}$, a pre-trained LLM $\pi_{\text{ref}}$ with parameter $\boldsymbol{\theta}_0$, learning rate $\alpha$.}
{
\For{$t = 1,2,\dots, T$}{
sample a minibatch $B$ of size $N$ from $D$;\\
$\boldsymbol{\theta}_t =\boldsymbol{\theta}_{t-1} \!\!-\alpha\nabla_{\boldsymbol{\theta}} \frac{1}{N}\sum^N_{i=1} \!\! \left[\left(\log\frac{\pi (y_i^{+} |x_i)}{\pi_{\text{ref}} (y_i^{+} |x_i)} \! - \!\log \frac{\pi (y_i^{-} |x_i)}{\pi_{\text{ref}} (y_i^{-} |x_i)}\right) \! - \! \lambda \perp\![\Delta_\pi( y_i^-,y_i^+;x_i)] -\frac{1}{2\beta }\right]^{2}$;
}
}
\Return $\boldsymbol{\theta}_{T}$.
\end{algorithm}

\section{Experiments}
\subsection{Setup}
\textbf{Datasets}. We evaluate the effectiveness of the proposed methods on three widely-used benchmark datasets:
(i) MT-Bench \citep{zheng2023judging}, which is a multi-turn question set on writing, roleplay, extraction, reasoning, math, coding, knowledge I (STEM), and knowledge II (humanities/social science).
(ii) Vicuna-Bench \citep{vicuna2023}, which is a single-turn question set on writing, roleplay, generic, fermi, counterfactual, coding, math, and knowledge.
(iii) Open-LLM leader-board \citep{open-llm-leaderboard}, which includes (a) commonsense reasoning: Arc \citep{clark2018think}, HellaSwag \citep{zellers2019hellaswag}, WinoGrande \citep{sakaguchi2021winogrande};
(b) multi-task language understanding: MMLU \citep{hendryckstest2021};
(c) human falsehood mimic: TruthfulQA \citep{lin2021truthfulqa}; and
(d) math problem solving: GSM8k \citep{cobbe2021training}. Table \ref{tbl:summary_llm} shows more information on the Hugging-Face Open LLM Leaderboard datasets. 

\begin{table*}[h]
\centering
\caption{Information on the Hugging-Face Open LLM Leaderboard datasets. }
\label{tbl:summary_llm}
\begin{tabular}{ccccccc}
\toprule
& Arc & TruthfulQA & Winogrande & GSM8k & HellaSwag & MMLU \\ 
\midrule
\# few-shot       & 25           & 0                   & 5                  & 5              & 10                & 5             \\ 
\midrule
performance metric            & acc\_norm    & mc2                 & acc                & acc            & acc\_norm         & acc           \\ 
\bottomrule
\end{tabular}
\end{table*}

\textbf{Setup for MT-Bench and Vicuna-Bench.} 
As in \citep{rafailov2023direct}, we use Pythia 2.8B \citep{biderman2023pythia}, a pretrained LLM without supervised fine-tuning and RLHF, as the backbone model. Following \citep{amini2024direct}, we first conduct supervised fine tuning (SFT) using the HH-RLHF dataset \citep{bai2022training}, which is human preference data on helpfulness and harmlessness based on positive feedbacks. We then perform direct alignment using the HH-RLHF dataset on the SFT model. Finally, we follow \citep{rafailov2023direct} to use GPT-4~\cite{achiam2023gpt} as a judge to evaluate the testing performance of the direct alignment trained model. For performance evaluation, we use the win-rate, tie-rate, and lose rate as in \citep{wang2023beyond}.

\textbf{Setup for Open-LLM Leader-board.}
Following \citep{tunstall2023zephyr,chen2024self}, we use zephyr-7b-sft-full\footnote{\url{https://huggingface.co/alignment-handbook/zephyr-7b-sft-full}}, a supervised fine-tuned version of Mistral 7B \citep{jiang2023mistral}, as the basic model. We directly perform direct alignment using a large-scale diverse preference dataset Ultra-feedback \citep{cui2023ultrafeedback}. The fine-tuned model is then evaluated on the testing benchmarks via the platform~\citep{eval-harness}.

Following \citep{wang2023beyond}, we use three performance metrics: 

i) Average marginal: $\frac{1}{N}\sum^N_{i=1}\log \frac{\pi\left(y_i^{+} \mid x_i\right)}{\pi_{\mathrm{ref}}\left(y_i^{+} \mid x_i\right)}-\log \frac{\pi\left(y_i^{-} \mid x_i\right)}{\pi_{\mathrm{ref}}\left(y_i^{-} \mid x_i\right)}$, which measures the gap between positive and negative responses.

(ii) Accuracy: $\frac{1}{N}
\sum^N_{i=1}\mathbbm{1} \left(\log \frac{\pi(y_i^{+} \mid x_i)}{\pi_{\mathrm{ref}}(y_i^{+} \mid x_i)}> \log \frac{\pi(y_i^{-} \mid x_i)}{\pi_{\mathrm{ref}}(y_i^{-} \mid x_i)}\right)$, 
which measures the number of tuples with the reward of a positive response larger than that of the negative response.
Here,
$\mathbbm{1}(\cdot)$ returns 1 when the argument holds and 0 otherwise. 

(iii) Accuracy defined on the prompt-augmented tuples: 
$\frac{1}{N}\sum^N_{i=1}\mathbbm{1} \left(\log \frac{\pi(y_i^{+} \mid p\oplus x_i)}{\pi_{\mathrm{ref}}(y_i^{+} \mid p\oplus x_i)}> \right.$ $\left.\log \frac{\pi(y_i^{-} \mid p\oplus x_i)}{\pi_{\mathrm{ref}}(y_i^{-} \mid 
p\oplus x_i)} \right)$,
which measures the accuracy for tuples with input augmented by the prompt.

\textbf{Baselines and Implmentation Details.}
We choose two widely-adopted direct alignment baselines: (i) DPO \citep{rafailov2023direct}, and (ii) IPO \citep{azar2024general}. Following \citep{rafailov2023direct}, $\beta$ is set to 0.1 for DPO and IPO. We also set $\beta=0.1$ for the proposed Sr-DPO and Sr-IPO for fair comparison. We select the optimal $\lambda$ from $\{0.1,0.3,0.5,1\}$ on the first 50 tuples from the HH-RLHF testing dataset. For both the HH-RLHF and UltraChat datasets, following \citep{rafailov2023direct}, the learning rate is $5 \times 10^{-7}$, the optimizer is RMSprop \citep{hinton2012neural}, the batch size is 64, and we also use gradient normalization to help training. The maximal input token size in training is 512. All experiments are performed on 8 A100 GPUs, and we use fully sharded data parallel~\citep{zhao2023pytorch} for distributed training.

\subsection{Performance Results}
\textbf{MT-Bench and Vicuna-Bench.} 
Table \ref{tbl:mt_vic} shows the testing win/tie/lose rates of the various methods as evaluated by GPT-4. As can be seen, the proposed Sr-DPO and Sr-IPO are effective and outperform DPO and IPO (in other words, the win rate is larger than the lose rate) in both MT-bench and Vicuna-bench. This reveals the effectiveness of our proposal methods. Some examples are also shown in Appendix \ref{appendix:case}.

\begin{table*}[h]
\centering
\caption{Testing performance on MT-Bench and Vicuna-Bench.}
\label{tbl:mt_vic}
\begin{tabular}{c|ccc|ccc}
\toprule
& \multicolumn{3}{c|}{MT-Bench} & \multicolumn{3}{c}{Vicuna-Bench} \\
& Win Rate & Tie Rate & Lose Rate &  Win Rate & Tie Rate & Lose Rate \\ 
\midrule
Sr-DPO vs DPO           & 45.62\%         & 33.76\%          & 20.62\%       & 63.75\%  &    13.75\%           &        22.50\%     \\ 
Sr-IPO vs IPO            & 38.75\%       & 25.00\%         & 26.25\%       & 60.00\% & 8.75\%   & 31.25\%             \\ 
Sr-DPO vs IPO              & 42.50\%       & 21.25\%          & 36.25\%       & 66.25\%       &     11.50\%     &         26.25\%    \\ 
Sr-IPO vs DPO             & 38.61\%         &    28.89\%    &    32.50\%     & 61.25\%       &   5.00\%        & 33.75\% \\ 
\bottomrule
\end{tabular}
\end{table*}

\begin{figure}[!h]
    \centering
    \begin{subfigure}{0.32\textwidth}
\includegraphics[width=\linewidth]{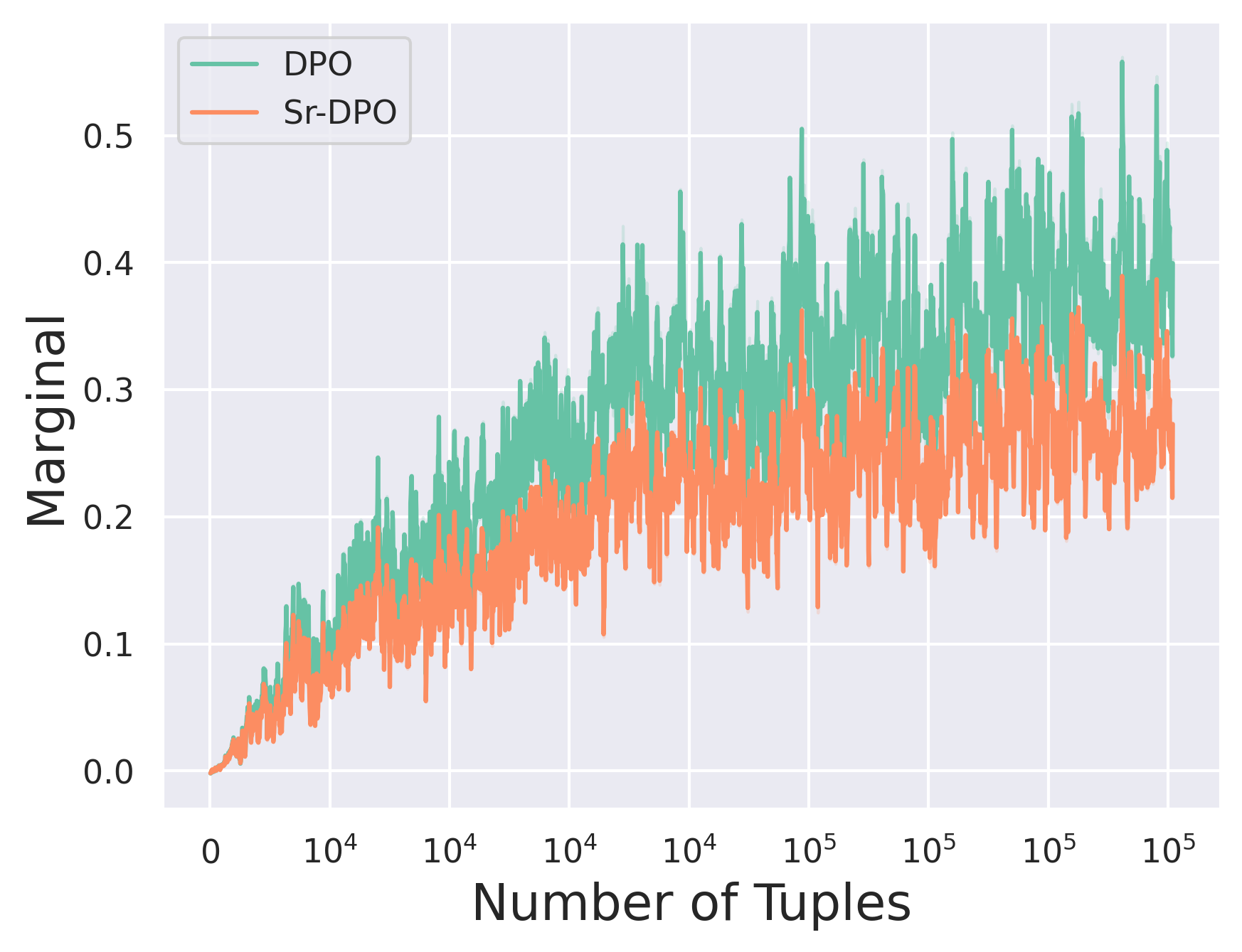} 
 \caption{Marginal: DPO vs Sr-DPO.}
\label{fig:dpo_vs_dpo_cali_mar}
    \end{subfigure}
        \hfill 
            \begin{subfigure}{0.32\textwidth}
\includegraphics[width=\linewidth]{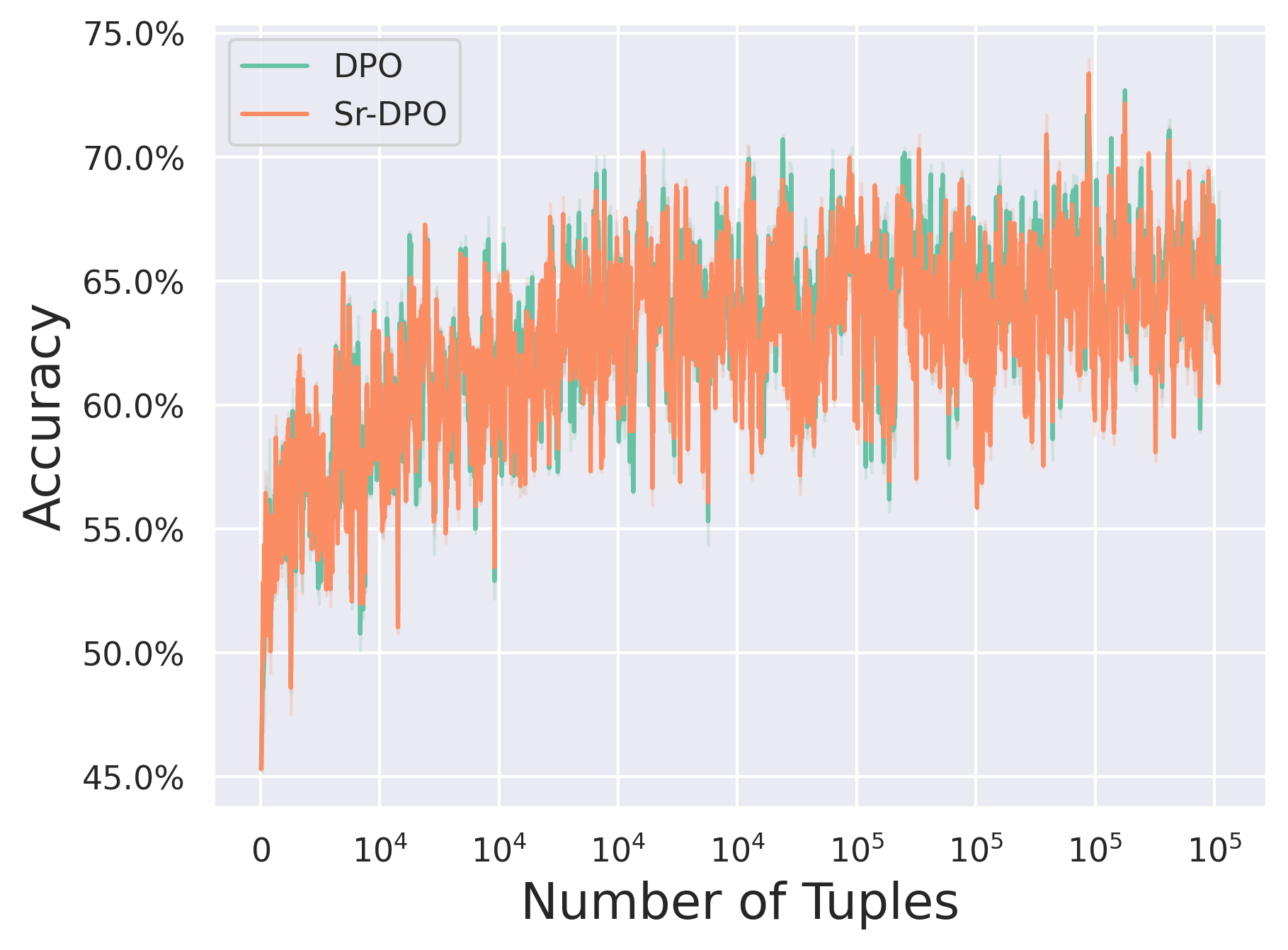}
\caption{Accuracy: IPO vs Sr-IPO.}
    \label{fig:dpo_vs_dpo_cali_acc}
    \end{subfigure}
            \hfill 
            \begin{subfigure}{0.32\textwidth}
\includegraphics[width=\linewidth]{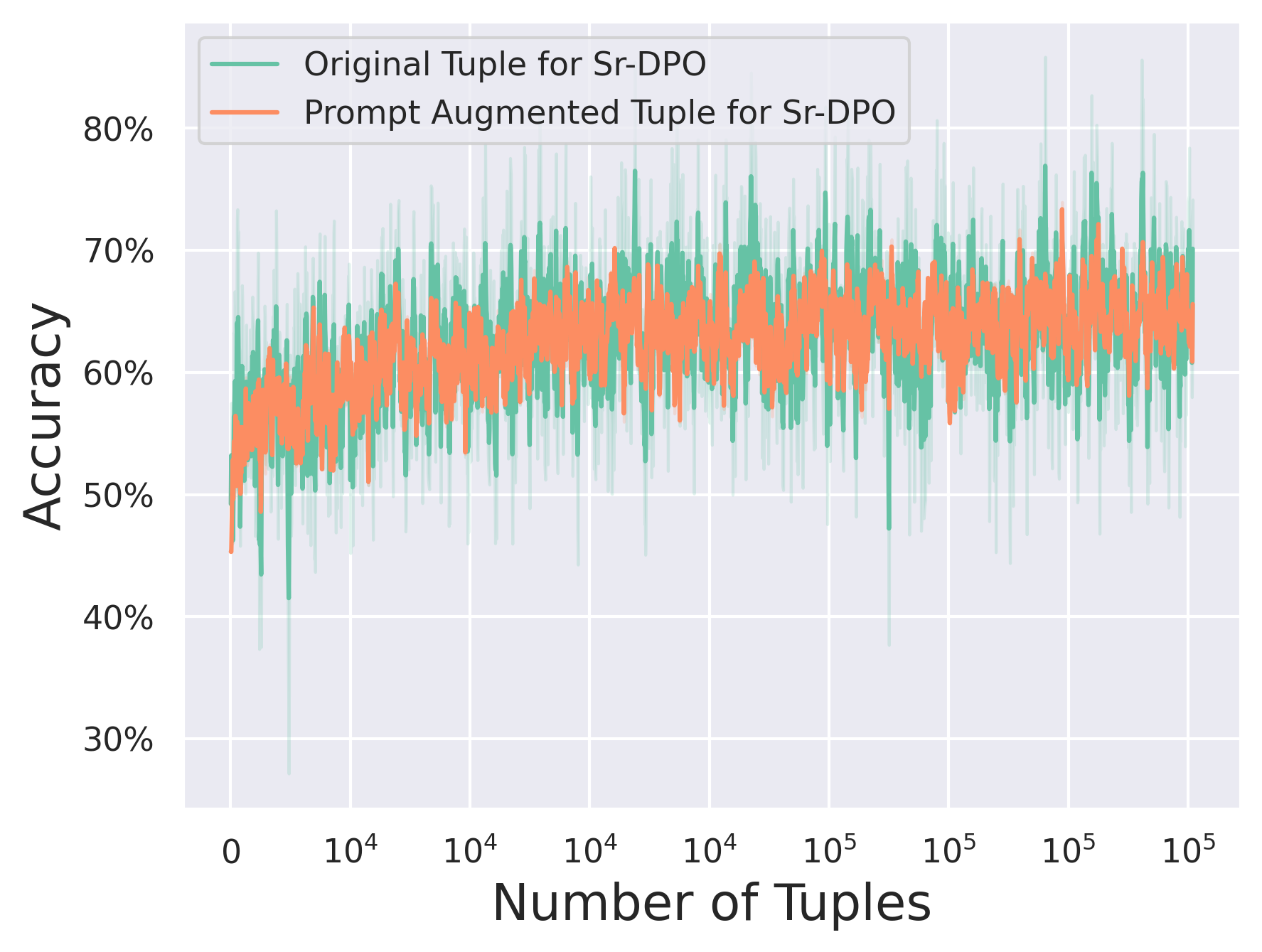}
\caption{Sr-DPO's accuracies for augmented and original tuples.}
    \label{fig:dpo_marg_delta}
    \end{subfigure}
        \hfill 
    \begin{subfigure}{0.32\textwidth}
\includegraphics[width=\linewidth]{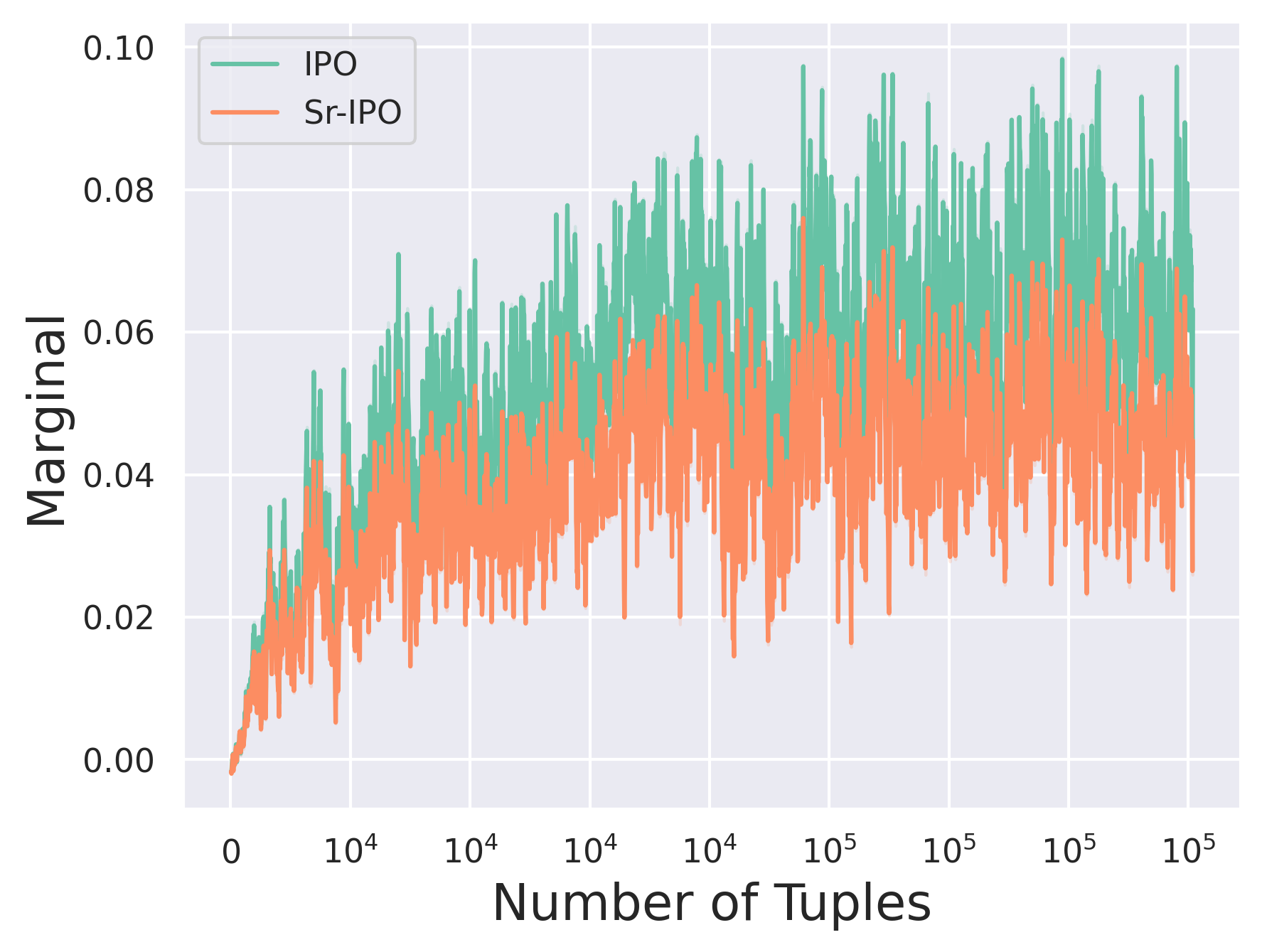}
\caption{Marginal: IPO vs Sr-IPO.}
    \label{fig:ipo_vs_ipo_cali_mar}
    \end{subfigure}
            \hfill 
                \begin{subfigure}{0.32\textwidth}
\includegraphics[width=\linewidth]{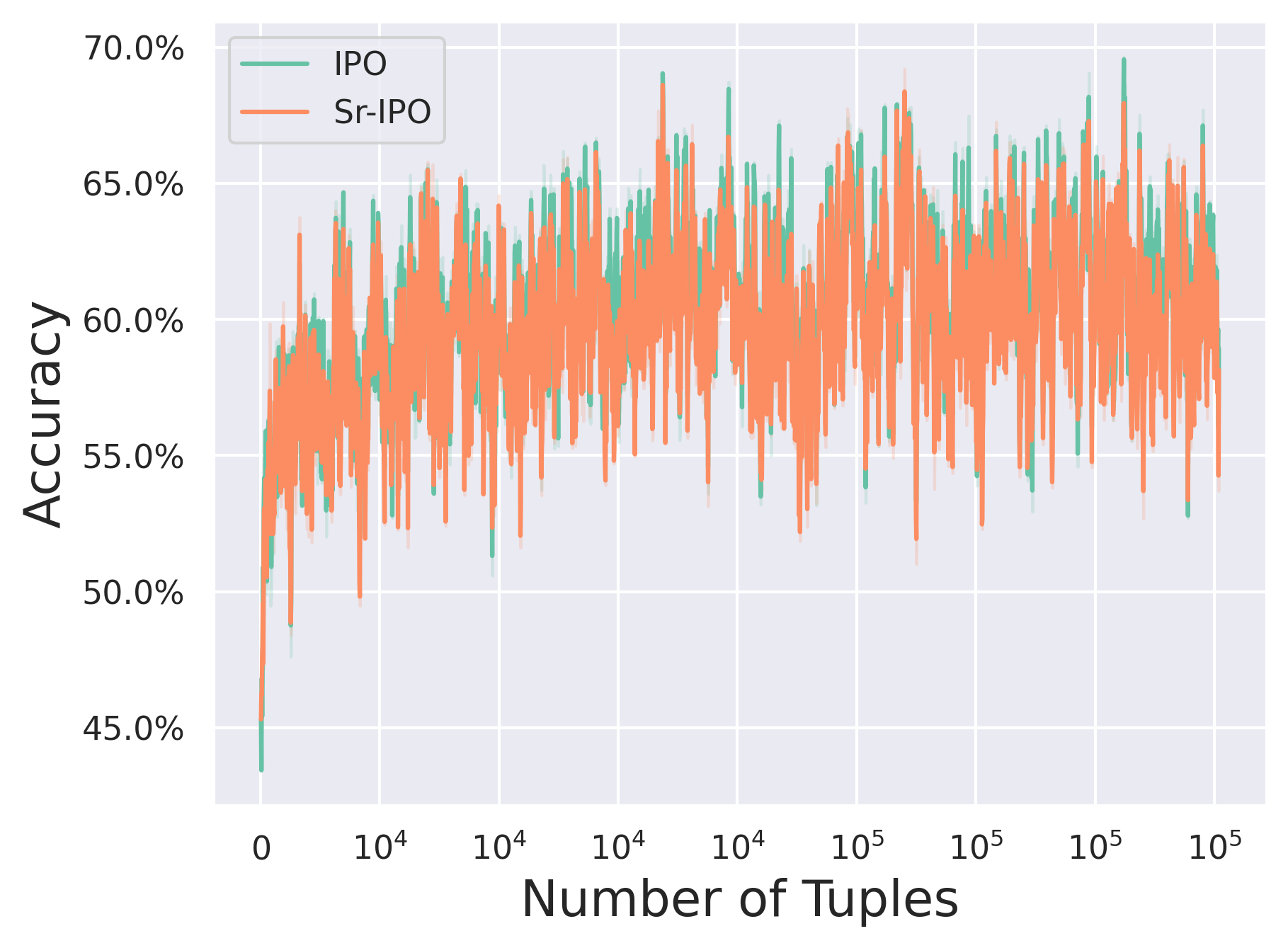}
\caption{Accuracy: IPO vs Sr-IPO.}
    \label{fig:ipo_vs_ipo_cali_acc}
    \end{subfigure}
            \hfill 
                \begin{subfigure}{0.32\textwidth}
\includegraphics[width=\linewidth]{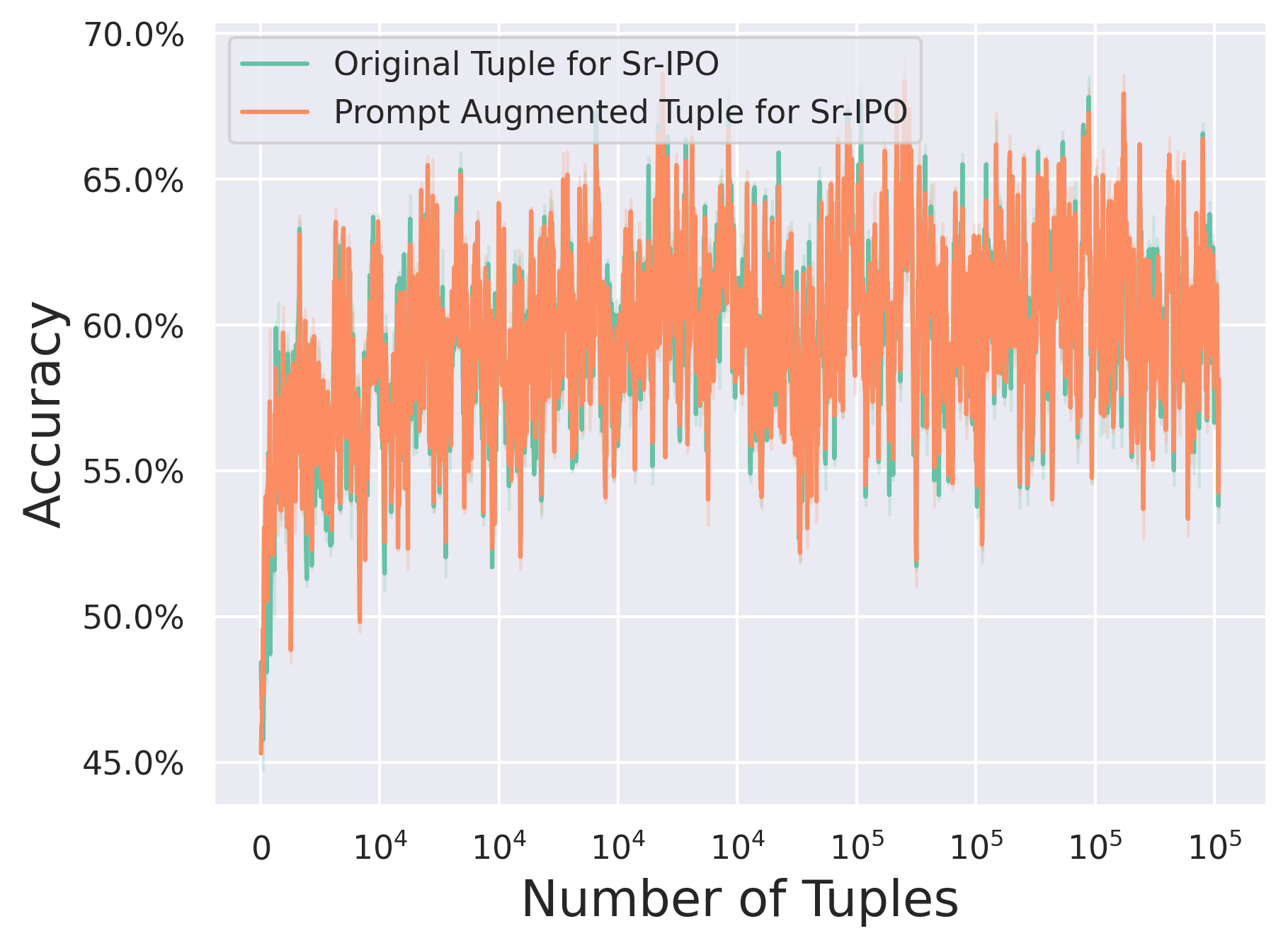}
\caption{Sr-IPO's accuracies for augmented and original tuples.}
    \label{fig:ipo_marg_delta}
    \end{subfigure}
\caption{Average marginal and accuracy on HH-RLHF dataset with different numbers of training tuples.}
    \label{fig:hh}
\end{figure}

\textbf{HH-RLHF.}
Figure \ref{fig:hh} shows results on the training set. As seen, the proposed Sr-DPO and Sr-IPO have lower marginals (Figures~\ref{fig:dpo_vs_dpo_cali_mar} and \ref{fig:ipo_vs_ipo_cali_mar}), while maintaining comparable accuracies (Figures \ref{fig:dpo_vs_dpo_cali_acc} and \ref{fig:ipo_vs_ipo_cali_acc}). This shows they avoid always enlarging the reward difference between positive and negative pairs, yet still achieving high accuracies. From Figures \ref{fig:dpo_marg_delta} and \ref{fig:ipo_marg_delta}, we find that accuracies for the prompt-augmented tuples are similar to those for the original tuples, thus verifying Assumption \ref{assu:consistency}.

\begin{table*}[!ht]
\centering
\small
\caption{Testing performance of the various methods on Open-LLM leader-board. The best one is in bold.}
\begin{tabular}{c|c|c|c|c|c|c|c}
\toprule
& Arc & TruthfulQA & WinoGrande & GSM8k & HellaSwag & MMLU & Average \\ \midrule
Zephyr-7b-sft-full & 60.41 & 43.73 & 74.19 & 26.76 &  82.85 &  60.92 & 58.14 \\ 
DPO & 63.05 & 59.30 & 78.53 & 31.01 &  84.89 &  61.44 & 63.04  \\
IPO & 61.86 & 46.03 & 79.24 & 35.91 & 82.80  & 61.55  & 61.23 \\ 
Sr-DPO & \textbf{65.27} & \textbf{60.30}  & \textbf{80.82} & 33.36  & \textbf{85.52}   & 61.63  &  \textbf{64.48}\\ 
Sr-IPO & 62.46 & 47.87 & 79.32 & \textbf{39.80}  & 83.15  &  \textbf{61.80} & 62.40 \\
\bottomrule
\end{tabular}
\label{tab:model_scores}
\end{table*}

\textbf{Open-LLM leader-board.} 
Table \ref{tab:model_scores} shows the testing performance across various methods. The results indicate that Sr-DPO achieves superior performance on Arc, TruthfulQA, WinoGrande; while Sr-IPO excels on GSM8k and MMLU. Overall, Sr-DPO emerges as the most effective method. Furthermore, Sr-DPO  (resp. Sr-IPO) consistently outperforms DPO (resp. IPO) across all six datasets. These results validate the effectiveness of the proposed approach.

\subsection{Effects of $\lambda$ and Number of Training Tuples}
\begin{wrapfigure}{l}{0.65\textwidth}
\vspace{-0.1in}
\begin{subfigure}{0.29\textwidth}
\includegraphics[width=\linewidth]{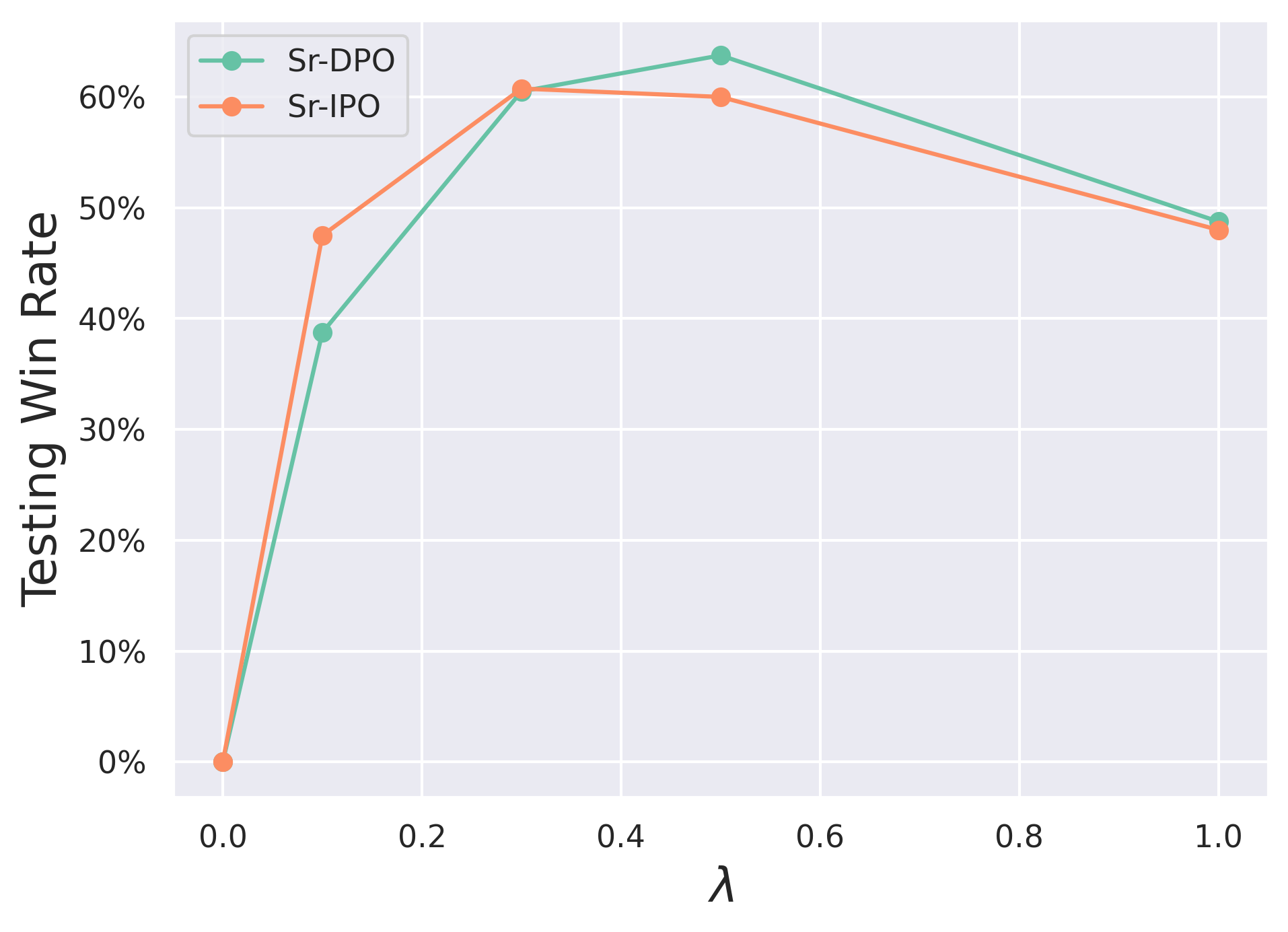}
\caption{Variation with $\lambda$.}
\label{fig:curve_win_rate}
\end{subfigure}
\hfill 
\begin{subfigure}{0.29\textwidth}
\includegraphics[width=\linewidth]{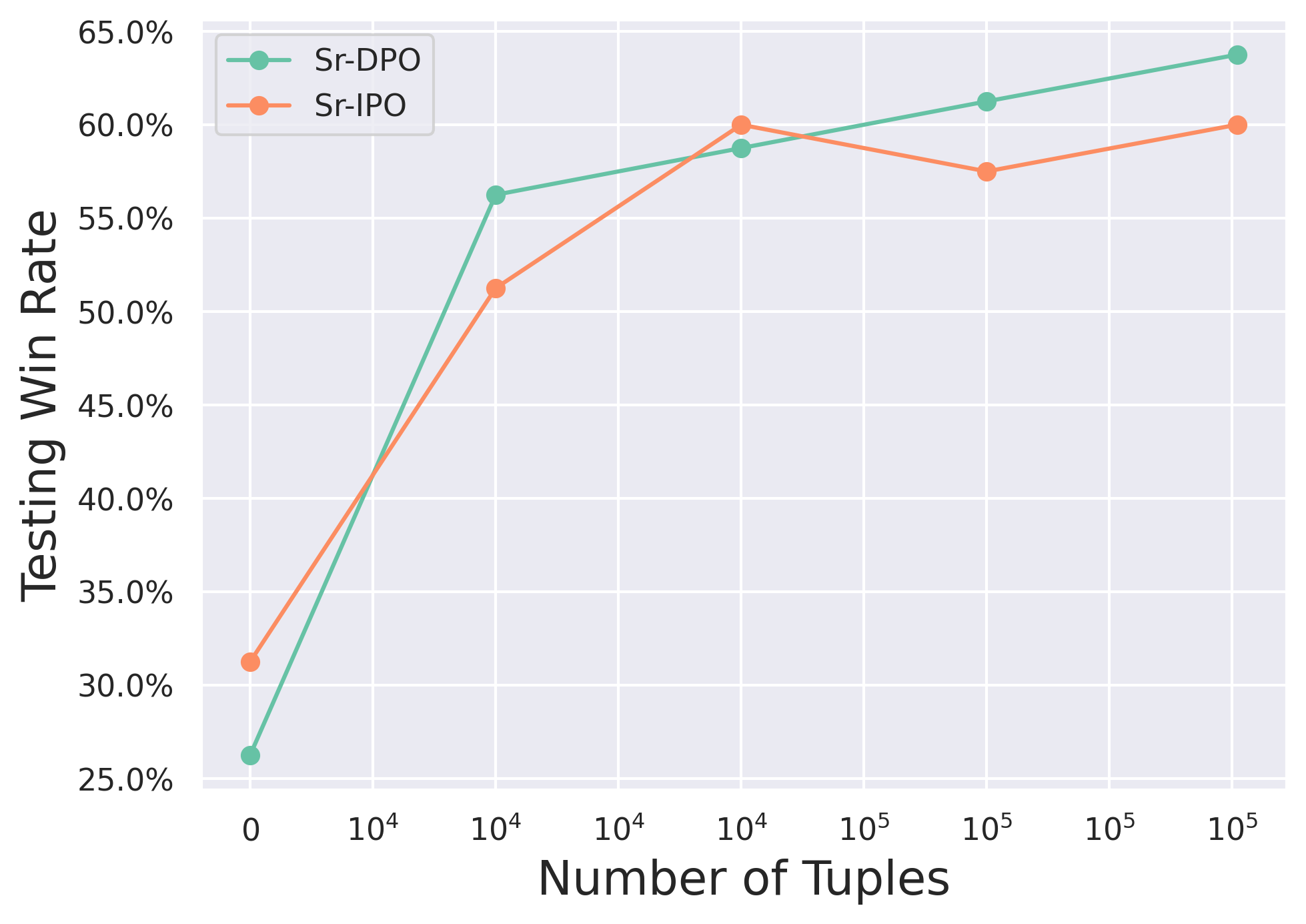} 
\caption{Variation with \#tuples.}
\label{fig:abs_win_rate}
\end{subfigure}
\caption{Win rates of Sr-DPO (vs DPO) and Sr-IPO (vs IPO) w.r.t. $\lambda$ and number of training tuples on Vicuna-Bench.}
\label{fig:step_metric}
\vspace{-0.1in}
\end{wrapfigure}

\textbf{Effect of $\lambda$.}
Figure \ref{fig:curve_win_rate} shows the testing win rate with varying $\lambda$ in Sr-DPO (resp. Sr-IPO) on MT-Bench and Vicuna-Bench. In both cases the win rate first increases with $\lambda$ and then decreases. In particular, $\lambda=0$ (i.e., not using the proposed refinement) leads to the worst performance. However, a $\lambda$ too large exaggerates the influence of $\Delta$, which can also negatively impact performance. We do not study its effect on Open-LLM leader-board because it is very time-consuming.

\textbf{Effect of the number of training tuples.}
Figure \ref{fig:abs_win_rate} shows the testing win rate with varying number of training tuples on Vicuna-Bench. As can be seen, both Sr-DPO and Sr-IPO can benefit from the use of more training tuples.

\begin{figure}[!ht]
    \centering
    \begin{subfigure}{0.32\textwidth}
\includegraphics[width=\linewidth]{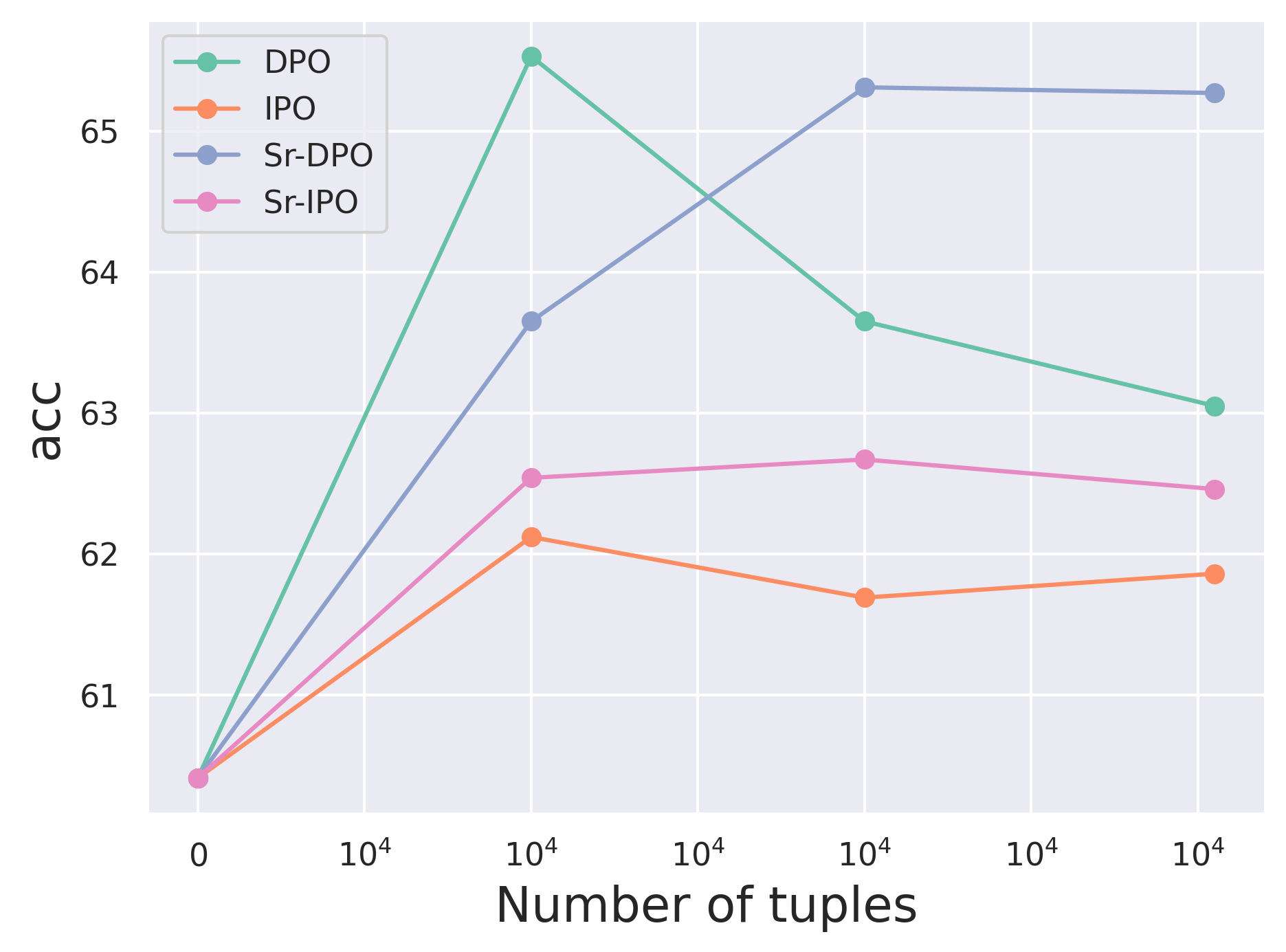} 
        \caption{ARC.}
\label{fig:arc_curve}
    \end{subfigure}
    \hfill 
    \begin{subfigure}{0.32\textwidth}
\includegraphics[width=\linewidth]{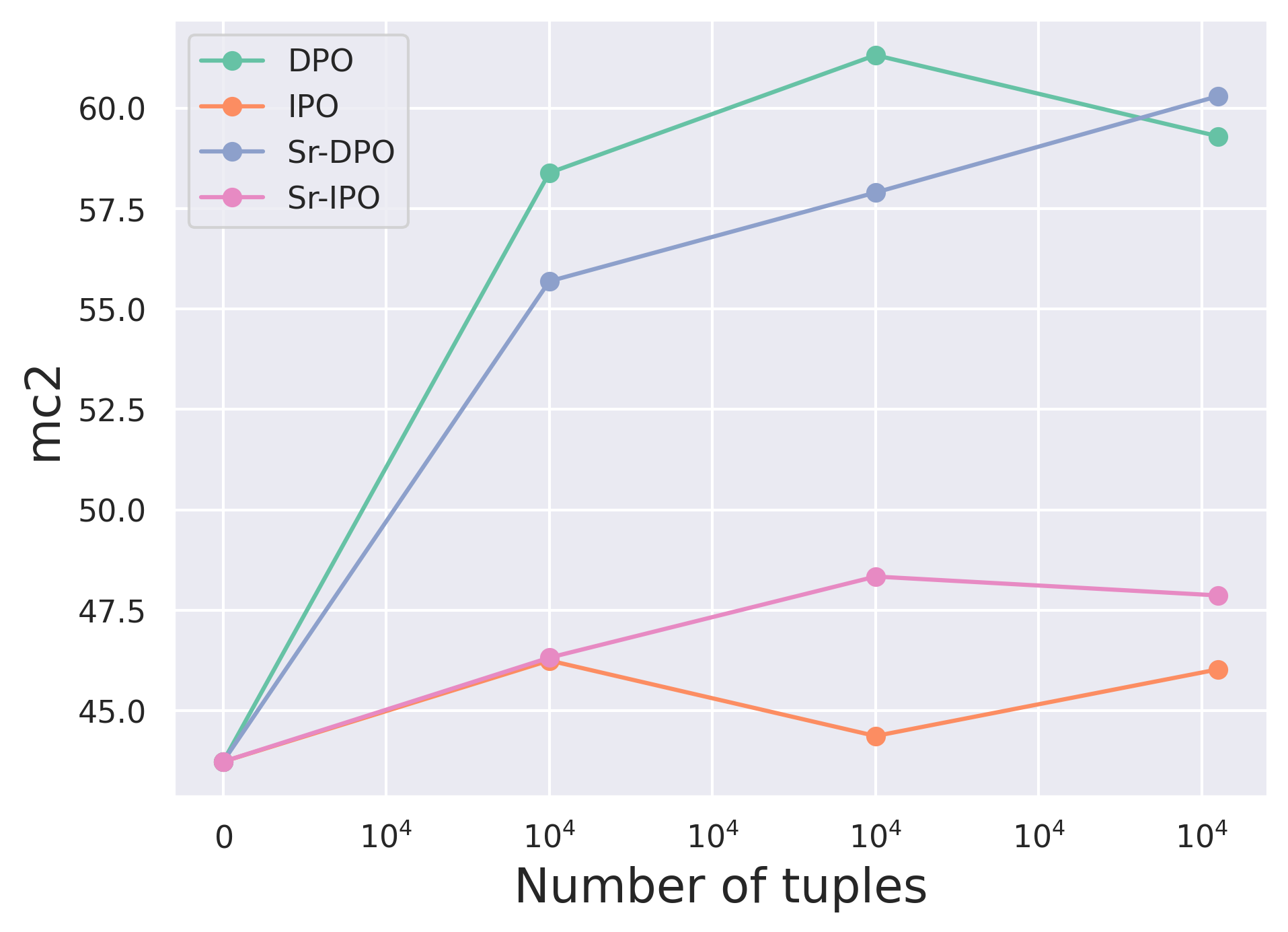}
                \caption{TruthfulQA.}
    \label{fig:trustfulqa_curve}
    \end{subfigure}
        \hfill 
        \begin{subfigure}{0.32\textwidth}
\includegraphics[width=\linewidth]{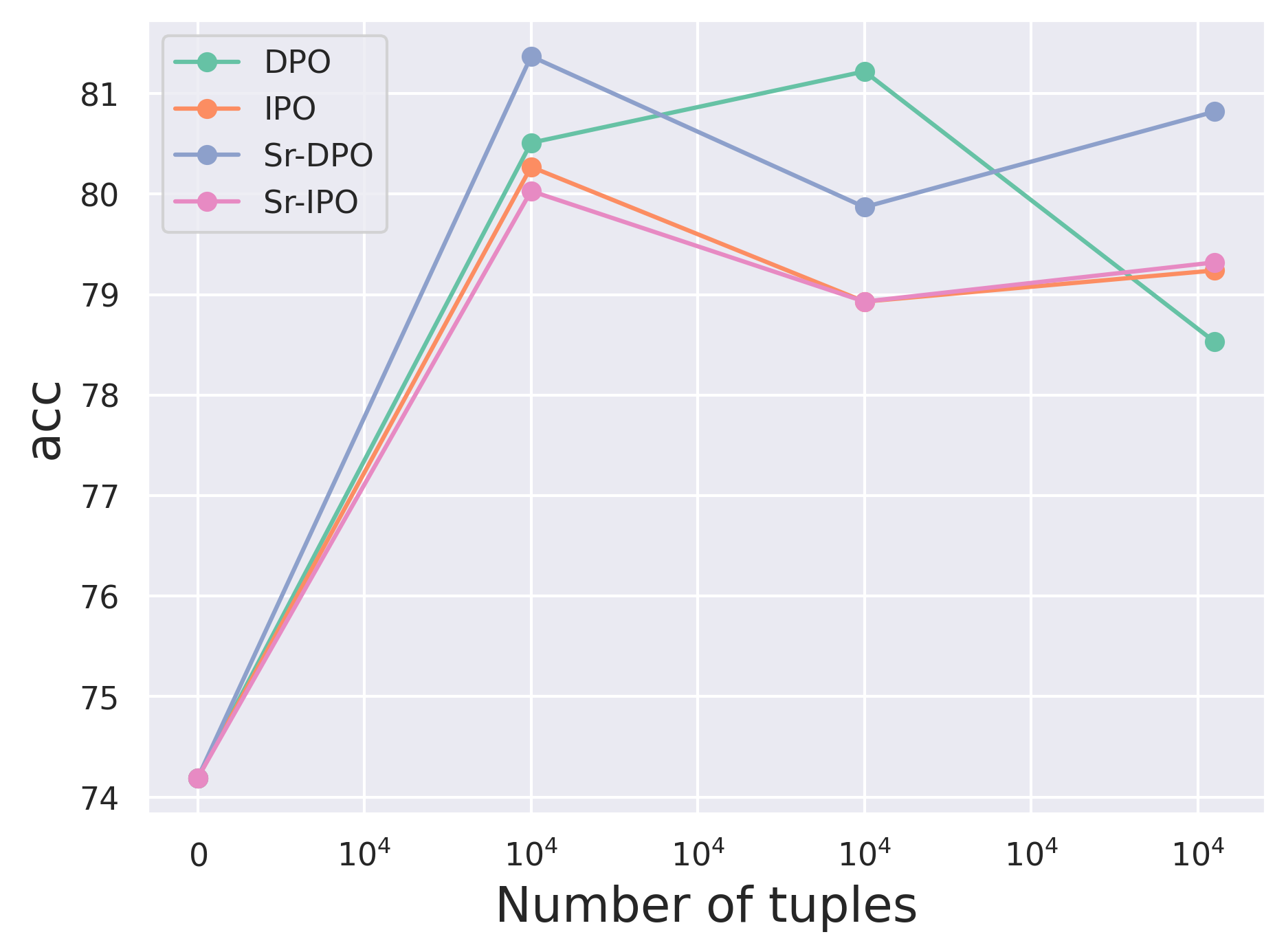}
        \caption{WinoGrande.}
\label{fig:WinoGrande_curve}
    \end{subfigure}
        \begin{subfigure}{0.32\textwidth}
\includegraphics[width=\linewidth]{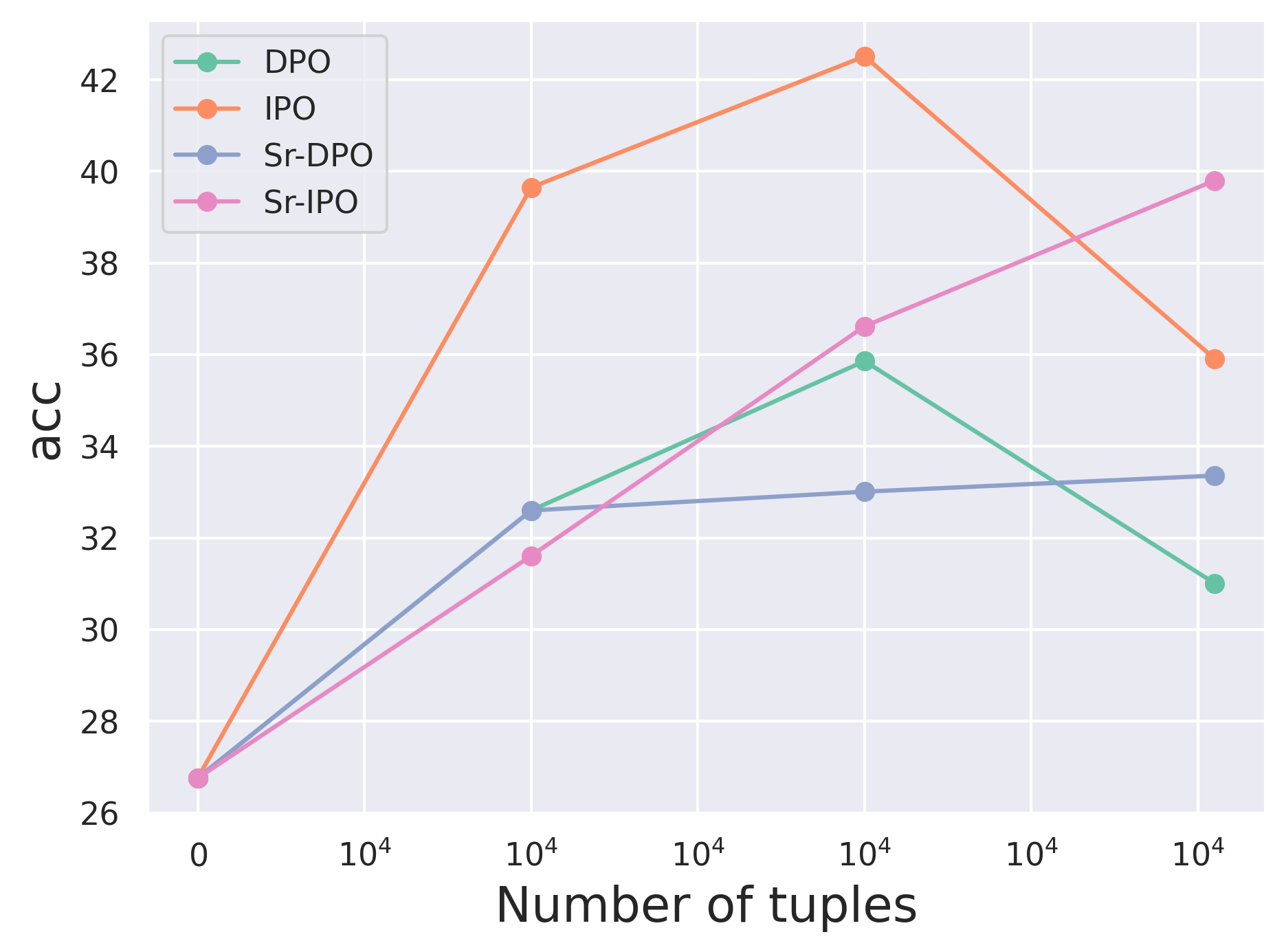} 
        \caption{GSM8k.}
    \end{subfigure}
    \hfill 
    \begin{subfigure}{0.32\textwidth}
\includegraphics[width=\linewidth]{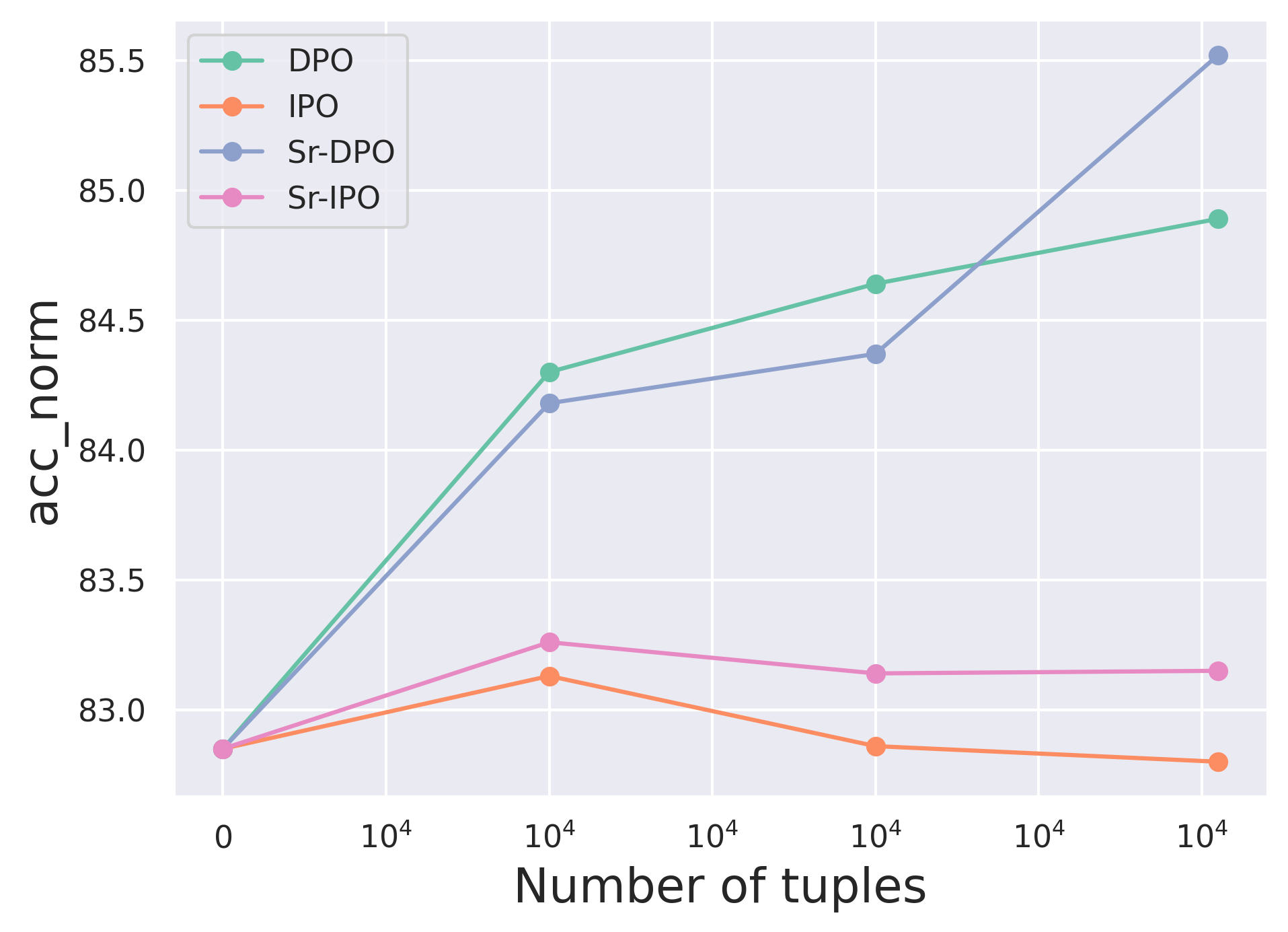}
                \caption{HellaSwag.}
    \end{subfigure}
        \hfill 
        \begin{subfigure}{0.32\textwidth}
\includegraphics[width=\linewidth]{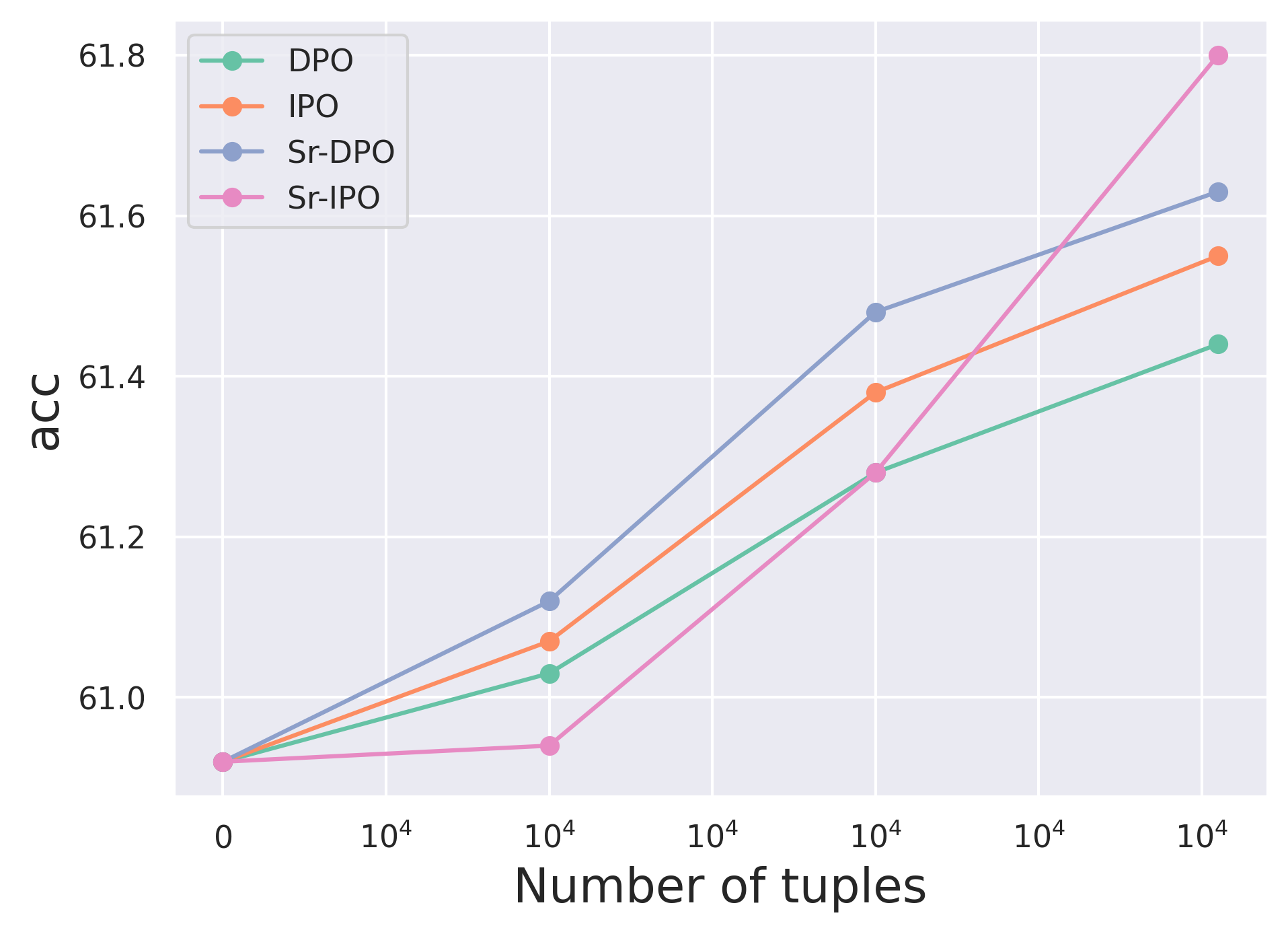}
        \caption{MMLU.}
    \end{subfigure}
    \caption{Performance  on 
Open-LLM leader-board
    with different numbers of training tuples.}
\label{fig:step_metric_leader}
\end{figure}

Figure~\ref{fig:step_metric_leader} shows the testing performance with varying number of training tuples on Open-LLM leader-board. As can be seen, DPO and IPO exhibit performance drop with the use of more training tuples on 4 of the 6 tasks (except on MMLU and HellaSwag), suggesting potential overfitting. In contrast, Sr-DPO and Sr-IPO are less prone to overfitting and exhibit improved performance with increased training data on ARC, TruthfulQA, GSM8k, and MMLU. 

\subsection{Correlation between Marginal and Score}
Recall that GPT-4 is used as the evaluator. For each instance $x_i$, we define the score difference between the corresponding positive $y_i^+$ and negative $y_i^-$ responses as $s^+(x_i) - s^-(x_i)$, where $s^+(x_i)$ (resp. $s^-(x_i)$) is GPT-4's evaluation score (0-5) of $y_i^+$ (resp. $y_i^-$). Inspired by \citep{kim2024prometheus}, which uses correlation to measure the agreements between GPT-4 and testing LLMs, here we compute the correlation between the marginal $\log \frac{\pi\left(y_i^{+} \mid x_i\right)}{\pi_{\mathrm{ref}}\left(y_i^{+} \mid x_i\right)} - \log \frac{\pi\left(y_i^{-} \mid x_i\right)}{\pi_{\mathrm{ref}}\left(y_i^{-} \mid x_i\right)}$ and score difference $s^+(x_i) - s^-(x_i)$ over 200 random tuples from the Ultrafeedback-binarized-preferences dataset\footnote{\url{https://huggingface.co/datasets/argilla/ultrafeedback-binarized-preferences}} (which already contains $s^+(x)$'s and $s^-(x)$'s). In particular, we use (i) Pearson’s correlation~\citep{sedgwick2012pearson}, which measures linear relationships; (ii) Spearman’s rank correlation~\citep{sedgwick2014spearman}, which measures monotonic relationships; and (iii) Kendall’s Tau~\citep{noether1981kendall}, which evaluates the strength and direction of associations.

Table \ref{tbl:corr} shows the correlation. As can be seen, Sr-DPO (resp. Sr-IPO) exhibits higher correlation values compared to DPO (resp. IPO). This suggests that Sr-DPO and Sr-IPO can evaluate the qualities of the positive and negative responses more accurately.

\subsection{Training Time}
Table \ref{tbl:time} shows the direct alignment training time on the HH-RLHF dataset. As can be seen, Sr-DPO and Sr-IPO are about $24\%$ slower than DPO and IPO. We consider this acceptable given the performance improvements achieved by Sr-DPO and Sr-IPO.

\begin{table*}[!htb]
\centering
\begin{minipage}{0.46\linewidth}
\centering
\captionof{table}{Correlation between the marginal and GPT-4's rating.}
\label{tbl:corr}
\begin{tabular}{@{}cccc@{}}
\toprule
 & Pearson's & Spearman's & Kendall’s \\
 & correlation & rank correlation& Tau \\
\midrule
DPO & 0.21 & 0.14 &  0.10 \\
IPO & 0.11 &  0.05& 0.03\\
Sr-DPO & 0.26 & 0.20 & 0.14 \\
Sr-IPO & 0.13 & 0.08 & 0.05 \\
\bottomrule
\end{tabular}
\end{minipage}%
\hspace{0.08\linewidth}
\begin{minipage}{0.45\linewidth}
\centering
\captionof{table}{Training time (in GPU hours) on HH-RLHF.}
\label{tbl:time}
\begin{tabular}{@{}cccc@{}}
\toprule
DPO & IPO & Sr-DPO & Sr-IPO \\ 
\midrule
16.8 & 16.8  & 20.8 & 20.8 \\
\bottomrule
\end{tabular}
\end{minipage}
\end{table*}

\vspace{-0.1in}
\section{Conclusion}
\vspace{-0.1in}
In this paper, we find that the widely adopted DPO method falls short by not accounting for the relative qualities of positive and negative samples, which can lead to sub-optimal training outcomes. To address this issue, we propose leveraging the intrinsic knowledge within LLMs to refine the loss function. Our main contributions are three-fold:
1) We utilize the knowledge of LLMs to create a refinement function that effectively estimates the quality of both positive and negative responses.
2) We demonstrate that under mild assumptions, the refinement function can enable the loss function to self-refine, leading to better alignment with human preferences.
3) Based on the refinement function, we develop two practical algorithms that enhance the training process.
Experimental results from various evaluators indicate that the proposed self-refined methods significantly improve the performance of fine-tuned models compared to existing approaches.

One limitation of our work is that we do not combine it with online policy-based direct alignment \citep{tang2024understanding}. We will investigate this in our future work.

\vspace{-0.1in}
\section{Broader Impacts}
\vspace{-0.1in}
Our work falls under one of the RLHF frameworks, and thus shares similar social impacts, both positive and negative. On the positive side, our method can help LLMs avoid generating harmful or unhelpful results by aligning with human feedback. However, conversely, our method also has the potential to enable LLMs to generate harmful or unhelpful results if it is used to align with malicious preferences.

\bibliographystyle{plain}
\bibliography{neurips}

\newpage
\appendix
\section{Proofs}
\label{sec:proofs}
\subsection{Proof of Proposition~\ref{prop:displacement}}
\begin{proof}
\noindent
(i) \underline{First property}: By directly substituting $\Delta_\pi(y^-, y^*; x)-\Delta_\pi(y^+, y^*; x) $ into the definition of $\Delta$ in (\ref{eq:delta}), we have
\begin{eqnarray*}
\Delta_\pi(y^-, y^*; x)-\Delta_\pi(y^+, y^*; x)  
&= & \beta \left( \log \frac{\pi(y^+ | p\oplus x)}{\pi_{0}(y^+ | p\oplus x)} - \log \frac{\pi(y^- | p\oplus x)}{\pi_{0}(y^- | p\oplus x)} \right) \\
&= & \beta \log \frac{\pi(y^+ | p\oplus x) \pi_{0}(y^- | p\oplus x)}{\pi(y^- | p\oplus x) \pi_{0}(y^+ | p\oplus x)} \\
& = & \Delta_\pi\left(y^- ,y^+;x \right).
\end{eqnarray*}

\noindent
(ii) \underline{Second property}: From assumption \ref{assu:improve}, we have: $ r\left( y^{+} |x\right)  >r\left( y^{-} |x\right), \forall (x,y^+,y-)$ with $y^+\succ y^-$. From (\ref{eq:dpo}),
\begin{equation*}
r( y^{+} |x)  >r\left( y^{-} |x\right)
\Leftrightarrow \log\frac{\pi \left( y^{+} \mid x\right)}{\pi_{0}\left( y^{+} \mid x\right)}  >\log\frac{\pi \left( y^{-} \mid x\right)}{\pi_{0}\left( y^{-} \mid x\right)}.
\end{equation*}

From Assumption \ref{assu:consistency}, for the prompt-augmented query $p\oplus x$, we still have $y^+ \succ y^-$. By using Assumption \ref{assu:improve} on $(p\oplus x,y^+,y^-)$, we have:

\begin{equation}\label{eq:key_step_theorem}
\displaystyle \log\frac{\pi (y^{+} |p\oplus x)}{\pi_{0} (y^{+} |p\oplus x)}  >\log \frac{\pi (y^{-} |p\oplus x)}{\pi_{0} (y^{-} |p\oplus x)}.
\end{equation}
Thus, from the first property,
\[ \Delta_\pi(y^+, y^*; x) - \Delta_\pi(y^-, y^*; x) 
= -\Delta_\pi\left(y^- ,y^+;x \right)
= -\beta \log \frac{\pi(y^+ | p\oplus x) \pi_{0}(y^- | p\oplus x)}{\pi_{0}(y^+ | p\oplus x) \pi(y^- | p\oplus x)} < 0 \]
because of (\ref{eq:key_step_theorem}), and so
$\Delta_\pi(y^+, y^*; x) < \Delta_\pi(y^-, y^*; x)$.
Thus, 
\begin{equation}\label{eq:res}
y^+ \succ y^-\Leftrightarrow \Delta_\pi\left( y^{+} ,y^*;x\right) < \Delta_\pi\left( y^{-}, y^*;x\right).
\end{equation}
Since $y^+\succ y^- \Leftrightarrow r^*(y^+|x)>r^*(y^-|x)$. We obtain that $r^*(y^+|x)>r^*(y^-|x) \Leftrightarrow \Delta_\pi\left( y^{+} ,y^*;x\right) < \Delta_\pi\left( y^{-}, y^*;x\right)$.
\end{proof}

\subsection{Proof of Corollary~\ref{cor:relative_displacement}}
\begin{proof}
Using property (ii) in Proposition~\ref{prop:displacement} on the assumption $r^{*}( y_i^+|x)> r^{*}( y_j^+|x)$ in Corollary~\ref{cor:relative_displacement}, we have
\begin{equation}
r^{*}( y_i^+|x)> r^{*}( y_j^+|x) \Leftrightarrow -\Delta_\pi(y_i^+, y^*; x) > -\Delta_\pi(y_j^+, y^*; x).
\end{equation}

Similarly, using this on the assumption in Corollary~\ref{cor:relative_displacement} that $r^{*}( y_i^-|x)<r^{*}( y_j^-|x)$, we have
\begin{equation}
-r^{*}( y_i^-|x)> -r^{*}( y_j^-|x) \Leftrightarrow \Delta_\pi(y_i^-, y^*; x) > \Delta_\pi(y_j^-, y^*; x).
\end{equation}

Adding these two together, we obtain 
\begin{eqnarray*}
\lefteqn{r^{*}( y_i^+|x) -r^{*}( y_i^-|x)  <r^{*}( y_j^+|x) -r^{*}( y_j^-|x)}\\
& 
\Leftrightarrow & \Delta_\pi(y_i^- ,y^*;x) -\Delta_\pi( y_i^+ ,y^*;x)  
> \Delta_\pi( y_j^-,y^*;x) -\Delta_\pi( y_j^+ ,y^*;x)\\
& 
\Leftrightarrow & \Delta_\pi( y_i^-, y_i^+;x)  >\Delta_\pi( y_j^-, y_j^+ ;x),
\end{eqnarray*}
due to property (i) of Proposition~\ref{prop:displacement}. 
\end{proof}

\subsection{Proof of Proposition \ref{prop:ipo_key}}\label{sec:ipo_cali}
As in \citep{amini2024direct,azar2024general}, the closed-form solution of (\ref{eq:ipo_Cali}) is:
\begin{equation*}
\pi (y^{+} |x)\propto \pi_{\text{ref}} (y^{+} |x)\exp\left( \beta ^{-1}\mathbb{E}_{y^{+}\sim \pi_{\text{ref}}(\cdot|x)} p(y^{+} \succ \pi_{\text{ref}} |x) +\lambda r(y^{+} | p\oplus x)\right).
\end{equation*}
Similarly for $y^-$, we have
\begin{equation*}
\pi (y^{-} |x)\propto \pi_{\text{ref}} (y^{-} |x)\exp\left( \beta ^{-1}\mathbb{E}_{y^{-}\sim \pi_{\text{ref}}(\cdot|x)} p(y^{-} \succ \pi_{\text{ref}} |x) +\lambda  r(y^{-} | p\oplus x)\right).
\end{equation*}
After simplification, we have:
\begin{eqnarray*}
\frac{\pi (y^{+} |x)}{\pi (y^{-} |x)}
& = & \frac{\pi_{\text{ref}} (y^{+} |x)}{\pi_{\text{ref}} (y^{-} |x)} \exp\left( \beta^{-1} [ \mathbb{E}_{y^{+}\sim \pi_{\text{ref}}(\cdot|x)} p(y^{+} \succ \pi_{\text{ref}} |x) - \mathbb{E}_{y^{-}\sim \pi_{\text{ref}}(\cdot|x)} p(y^{-} \succ \pi_{\text{ref}} |x) ] \right. \\
&& \left. + \lambda [ r(y^{+} | p\oplus x) - r(y^{-} | p\oplus x) ] \right).
\end{eqnarray*}
Note that $r(y^{+} | p\oplus x) - r(y^{-} | p\oplus x) = \Delta_{\pi}( y^-,y^+;x)$. After some simplifications, we have
\begin{eqnarray*}
\lefteqn{\log\frac{\pi (y^{+} |x)}{\pi_{\text{ref}} (y^{+} |x)} -\log\frac{\pi (y^{-} |x)}{\pi_{\text{ref}} (y^{-} |x)}  }\\
 & =  & \frac{1}{\beta } \left[ \mathbb{E}_{y^{+}\sim \pi_{\text{ref}}(\cdot|x)} p(y^{+} \succ \pi_{\text{ref}} |x) - \mathbb{E}_{y^{-}\sim \pi_{\text{ref}}(\cdot|x)} p(y^{-} \succ \pi_{\text{ref}} |x)\right]+\lambda \Delta_{\pi}( y^-,y^+;x) .
\end{eqnarray*}
As $\mathbb{E}_{y^{+}\sim \pi_{\text{ref}}(\cdot|x)} p(y^{+} \succ \pi_{\text{ref}} |x) - \mathbb{E}_{y^{-}\sim \pi_{\text{ref}}(\cdot|x)} p(y^{-} \succ \pi_{\text{ref}} |x)= \frac{1}{2}$ \citep{azar2024general}, after taking expectations of both sides, we obtain

\begin{eqnarray}\label{eq:ipo_key_eq}
\log\frac{\pi (y^{+} |x)}{\pi_{\text{ref}} (y^{+} |x)} -\log\frac{\pi (y^{-} |x)}{\pi_{\text{ref}} (y^{-} |x)} =  \frac{1}{2\beta}+\lambda \Delta_{\pi}( y^-,y^+;x) .
\end{eqnarray}

\section{Example Cases}\label{appendix:case}
Tables \ref{tab:case1}, \ref{tab:case2}, \ref{tab:case3}, and \ref{tab:case4} show four examples of responses by Sr-DPO, Sr-IPO, DPO, and IPO. As can be seen, the proposed Sr-DPO and Sr-IPO outperform DPO and IPO under GPT4's judgement.

\newpage

\begin{table*}[h!]
\centering
\caption{Example 1 for Sr-DPO and DPO. The first row is the query, the second and third rows are the responses generated by Sr-DPO and DPO, respectively, and the last row is the judgement made by GPT-4.}
\label{tab:case1}
\begin{tabular}{p{2cm}p{10cm}}
\toprule
\textbf{Prompt} & Explain the process of natural selection and how it contributes to the evolution and adaptation of species.\\
\midrule
\textbf{Sr-DPO} & 
Natural selection is the process of survival of the fittest, which occurs when organisms with the most advantageous traits are more likely to reproduce and pass on their traits than organisms with less desirable traits. It is a key factor in the evolution of species, with the most successful species being those that are the most adaptable to the environment. Natural selection ensures that organisms with the most advantageous traits are more likely to survive and pass on their traits to the next generation.

Here is a brief explanation of the process of natural selection:

1. The most adaptable organisms with the most advantageous traits are more likely to survive and pass on their traits to the next generation.

2. The most advantageous traits of organisms are those that are most effective at surviving and reproducing in their environment.

3. Those organisms with the most advantageous traits are more likely to survive and reproduce than organisms with less advantageous traits.
\\
\midrule
\textbf{DPO} & Natural selection is the process of survival of the fittest, which occurs when the genetic makeup of a population changes over time, favoring individuals that have beneficial traits such as increased physical strength, increased intelligence, or other adaptive features. This process allows for the survival of the species that is best able to adapt to its environment.\\

\midrule
\textbf{Judgment} & 
Both assistants provide a correct explanation of natural selection, describing it as a process that favors individuals with beneficial traits, leading to changes in the genetic makeup of a population over time. However, Assistant A's response is more detailed and comprehensive. It explains how the probability of an individual's survival and reproduction is influenced by how well adapted the organism is to its environment, and it also mentions the importance of natural selection in the field of biology and evolution. Therefore, Assistant A's response is more informative and provides a deeper understanding of the topic. 

Final Verdict: [[A]]
\\
\bottomrule
\end{tabular}
\end{table*}

\begin{table*}[h!]
\centering
\caption{Example 2 for Sr-DPO and DPO. The first row is the query, the second and third rows are the responses generated by Sr-DPO and DPO, respectively, and the last row is the judgement made by GPT-4. }
\label{tab:case2}
\begin{tabular}{p{2cm}p{10cm}}
\midrule
\textbf{Prompt} & Draft an apology email to a customer who experienced a delay in their order, and provide reassurance that the issue has been resolved.\\
\midrule
\textbf{Sr-DPO} & 
Here's an apology email draft for a customer who experienced a delay in their order:

Dear Customer,

Thank you for your patience during our recent order process.  We recently experienced a delay with your order and apologize for the inconvenience.  We are currently working to process your order, and we will provide a notification when it is ready.  We sincerely apologize for the delay and appreciate your patience.  Our team is working hard to resolve this issue as soon as possible.  We look forward to welcoming your order and providing you with the best customer service experience possible.

Best regards,

Customer Support Team
\\
\midrule
\textbf{DPO} & Here's an apology email draft for a customer who experienced a delay in their order:

Dear [Customer Name],

apologize for the delay in your delivery of your order. I have worked with our team to resolve the issue, and I am very pleased to let you know that the issue has been resolved and your order is now ready for dispatch. I look forward to providing you with a great customer experience in the future.

Sincerely,

[Customer Name]

Customer Support\\
\midrule
\textbf{Judgment} & 
Both assistants provided a good response to the user's request. However, Assistant A's response was more detailed and provided more reassurance to the customer. Assistant A mentioned that they reviewed the issue with their fulfillment partners and are now shipping the order as quickly as possible. This gives the customer a sense of urgency and care for their order. On the other hand, Assistant B's response was also good but lacked the level of detail and reassurance that Assistant A's response provided. Therefore, Assistant A's response is better. 

Final Verdict: [[A]]
\\
\bottomrule
\end{tabular}
\end{table*}

\begin{table*}[h!]
\centering
\caption{Example 3 for Sr-IPO and IPO. The first row is the query, the second and third rows are the responses generated by Sr-IPO and IPO, respectively, and the last row is the judgement made by GPT-4. }
\label{tab:case3}
\begin{tabular}{p{2cm}p{10cm}}
\toprule
\textbf{Prompt} & Why might someone choose to use a paper map or ask for directions instead of relying on a GPS device or smartphone app?\\
\midrule
\textbf{Sr-IPO} & 
A paper map or paper printed directions can be a fun way to help you get from point A to point B. Paper maps are portable and easy to use, and they're also a fun and easy way to explore your surroundings and get around town. They're also less expensive than GPS systems or smartphone apps, and they're very easy to use, so they're a good option for older kids or adults who need to get by without a smartphone. Paper maps offer a more personal and hands-on experience than smartphone apps or GPS devices, and they can be a fun way to get around town or explore your surroundings.
\\
\midrule
\textbf{IPO} & A paper map or paper printed directions can be a fun way to help you get from point A to point B. Paper maps are portable and easy to use, and they're also a fun and easy way to explore your surroundings and get around town. They're also less expensive than GPS systems or smartphone apps, and they're very easy to carry around.
If you are interested in using a paper map or directions, here are a few reasons why you might choose to do so:1. Paper maps are portable and easy to use.

2. Paper maps are less expensive than GPS systems or smartphone apps.

3. Paper maps are easy to carry around.

4. Paper maps are easy to use and interpret.

5. Paper maps can help you explore your surroundings and get around town.

6. Paper maps are easily outdated or outdated, so it is important to be sure you have the latest version of the map to use.

7. Paper maps are a fun way to explore your surroundings and get around town.

8. Paper maps are easy to use and interpret.

9. Paper maps can help you explore your surroundings and get around town.

10. Paper maps make it easy to get around town and explore your surroundings.

11. Paper maps are a fun and easy way to explore your surroundings and get around town.

Remember to tailor the activities to the birthday childs interests and preferences. Have a great celebration!\\
\midrule
\textbf{Judgment} & 
Assistant A's response is more concise and to the point, providing a clear explanation of why someone might choose to use a paper map or ask for directions instead of relying on a GPS device or smartphone app. Assistant B's response, while it does provide some valid points, is repetitive and confusing, especially the last sentence about tailoring activities to a birthday child's interests and preferences, which is completely irrelevant to the question asked. Therefore, Assistant A's response is more helpful and relevant. 

Final Verdict: [[A]]
\\
\bottomrule
\end{tabular}
\end{table*}

\begin{table*}[h!]
\centering
\caption{Example 4 for Sr-IPO and IPO. The first row is the query, the second and third rows are the responses generated by Sr-IPO and IPO, respectively, and the last row is the judgement made by GPT-4. }
\label{tab:case4}
\begin{tabular}{p{2cm}p{10cm}}
\toprule
\textbf{Prompt} & How many lightning strikes occur on Earth each day? Try to explain your answer. Your explanation should take the reader through your reasoning step-by-step.\\
\midrule
\textbf{Sr-IPO} & 
There are about 2,500 lightning strikes each day on Earth. That's about two lightning strikes for every minute! The most common type of lightning strikes occur in the tropics, where temperatures range from 70 to 90 degrees Fahrenheit. Lightning strikes are most common around noon and also around sunset. The next most common type of lightning occurs in the polar regions, where the temperatures can reach as low as -20 degrees Celsius. Lightning strikes are most common during the winter months in these regions.

The most dangerous type of lightning occurs when the atmospheric pressure is extremely low, like when a storm is approaching. Low atmospheric pressure can cause the air to move faster and more violently, creating a stronger electric current through the air. When the air is moving quickly, it is also possible for an electrical charge to build up between the tips of a cloud or a thunderstorm. This electrical charge can then be released as lightning. Low atmospheric pressure can also increase the intensity of a thunderstorm, leading to more severe lightning storms.
\\
\midrule
\textbf{IPO} & There are about 2,500 lightning strikes each day on Earth. That's about two lightning strikes for every minute! The most common type of lightning strikes occur in the tropics, where temperatures range from 70 to 90 degrees Fahrenheit. Lightning strikes are most common around noon and also around sunset. The next most common type of lightning occurs in the polar regions, where the temperatures can reach as low as -20 degrees Celsius. Lightning strikes are most common during the winter months in these regions.\\
\midrule
\textbf{Judgment} & 
Both Assistant A and Assistant B start their responses with the same information, stating that there are about 2,500 lightning strikes each day on Earth. They both also provide similar information about where and when lightning strikes are most common. However, Assistant A goes a step further by explaining the conditions under which the most dangerous type of lightning occurs, providing a more detailed and comprehensive response to the user's question. Therefore, Assistant A's response is more informative and thorough.

Final Verdict: [[A]]
\\
\bottomrule
\end{tabular}
\end{table*}

\end{document}